%% file: main.tex
\pdfoutput=1
\documentclass[10pt,twocolumn,letterpaper]{article}

\usepackage[preprint]{cvpr}      

\input{preamble}

\def\paperID{8654} 
\def\confName{CVPR}
\def\confYear{2024}

\usepackage{todonotes}
\usepackage[utf8]{inputenc} 
\usepackage[T1]{fontenc}    
\usepackage{hyperref}       
\usepackage{url}            
\usepackage{booktabs}       
\usepackage{amsfonts}       
\usepackage{nicefrac}       
\usepackage{microtype}      
\usepackage{xcolor}         
\usepackage{graphicx}
\usepackage{subcaption}
\usepackage{array}
\usepackage{caption}
\usepackage{natbib}
\setcitestyle{numbers, sort}
\usepackage{lipsum}
\usepackage{sidecap}
\usepackage{multirow}
\usepackage{makecell}
\usepackage{soul}
\usepackage{bm}
\newsavebox{\measurebox}

\newcommand{\abbrv}{BRICS}
\makeatletter
\newcommand{\thickhline}{%
    \noalign {\ifnum 0=`}\fi \hrule height 1pt
    \futurelet \reserved@a \@xhline
}
\newcolumntype{"}{@{\hskip\tabcolsep\vrule width 1pt\hskip\tabcolsep}}
\makeatother
\input{ourcommands}

\title{BRICS: Bi-level feature Representation of Image CollectionS}

\author{%
  Dingdong Yang$^1$ \quad Yizhi Wang$^1$ \quad Ali Mahdavi-Amiri$^1$ \quad Hao Zhang$^1$\thanks{Corresponding Author.} \\ \\
  $^1$School of Computing Science, Simon Fraser University
}

\begin{document}

\maketitle

\input{00_abstract}

\input{01_intro}

\input{02_RW}

\input{03_method}

\input{04_results}

\input{05_conclusion}

\clearpage
{
    \small
    \bibliographystyle{ieeenat_fullname}
    \bibliography{main}
}

\include{supp.tex}

\end{document}

%% file: preamble.tex
\usepackage[dvipsnames]{xcolor}



%% file: ourcommands.tex
\usepackage{tikz}
\newcommand*\circled[1]{\tikz[baseline=(char.base)]{
            \node[shape=circle,draw,inner sep=0.7pt] (char) {#1};}}

\newcommand{\am}[1]{\textcolor{orange}{#1}}

\newcommand\newsubcap[1]{\phantomcaption%
       \caption*{\figurename~\thefigure(\thesubfigure): #1}}

%% file: 00_abstract.tex
\begin{abstract}
We present \abbrv, a {\em{bi-level}\/} feature representation for image collections, which consists of a key code space on top of a feature grid space. Specifically, our representation is learned by an autoencoder to encode images into {\em continuous\/} key codes, which are used to retrieve features from {\em groups of multi-resolution feature grids\/}.
Our key codes and feature grids are jointly trained continuously with well-defined gradient flows, leading to high usage rates of the feature grids and improved generative modeling compared to discrete Vector Quantization (VQ).
Differently from existing continuous representations such as KL-regularized latent codes, our key codes are strictly bounded in 
scale and variance. Overall, feature encoding by \abbrv\ is compact, efficient to train, and enables generative modeling over key codes using the diffusion model. Experimental results show that our method achieves comparable reconstruction results to VQ while having a smaller and more efficient decoder network (\textbf{$\approx$50\% fewer} GFlops). By applying the diffusion model over our key code space, we achieve
state-of-the-art performance on image synthesis on the FFHQ and LSUN-Church (\textbf{$\approx$29\% lower} than LDM~\cite{rombach2022high}, \textbf{$\approx$32\% lower} than StyleGAN2~\cite{karras2020analyzing}, \textbf{$\approx$44\% lower} than Projected GAN~\cite{sauer2021projected} on CLIP-FID) datasets. 

\end{abstract}

%% file: 01_intro.tex
\section{Introduction}
\label{sec:intro}

\begin{figure}[t!]
\centering
\includegraphics[width=0.99\linewidth]{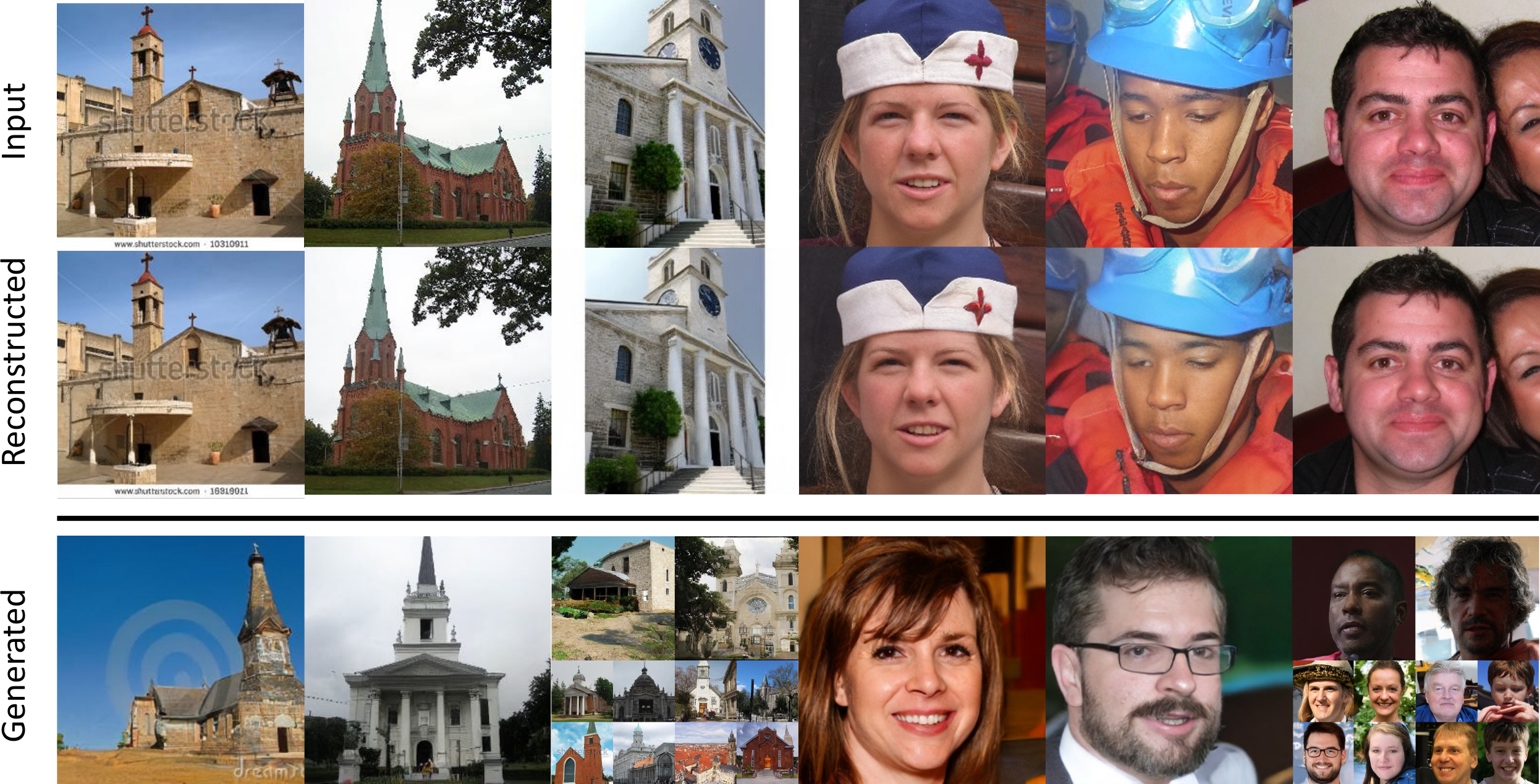}
\caption{Image reconstruction results by \abbrv\ (row 2) on the validation set of LSUN-church $256 \times 256$~\cite{yu2015lsun} and FFHQ $256 \times 256$ datasets~\cite{karras2019style}, where we note the recovery of fine details such as text captions and watermarks. Bottom row shows images generated via the diffusion model trained on key codes from \abbrv (zoom in to see more details)}.
\label{fig:best_showoff}
\end{figure}

\begin{figure}[t!]
\centering
\includegraphics[width=0.99\linewidth]{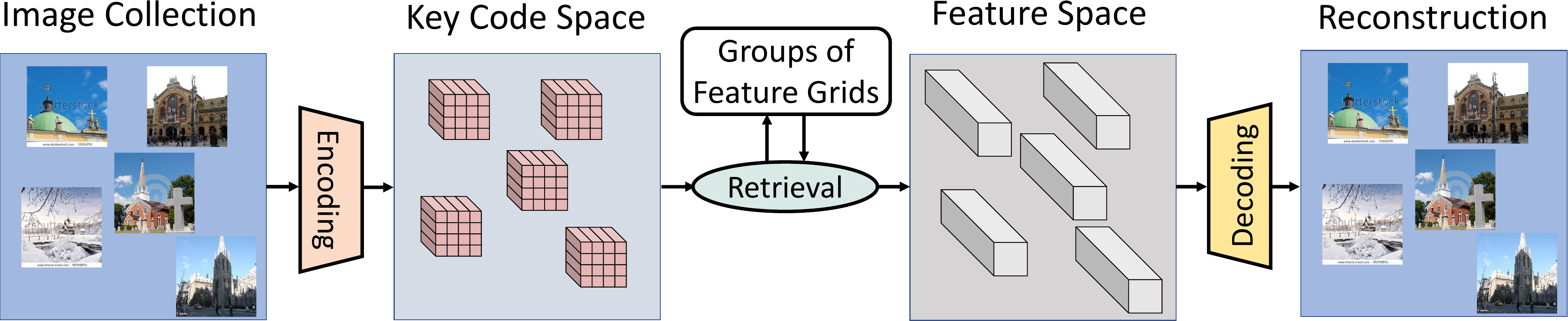}
\caption{Rather than directly encoding images into features, our method first projects images into key codes and then uses the key codes to retrieve features from groups of feature grids. The encoder, decoder, and feature grids are jointly learned via autoencoding.}
\label{fig:concept_of_method}
\end{figure}

One of the most foundational building blocks of modern-day neural information processing is feature representation learning,
whose goal is to map input data into a latent space with desirable properties to uncover useful data patterns and assist a 
task at hand. Commonly favoured properties of the latent representations include continuity, efficient access, 
interpretability, compactness while retaining reconstruction quality, as well as generative capabilities as reflected by the 
plausibility and diversity of the generated contents. Classical examples of representation learning include dictionary learning,
manifold learning, and autoencoders~\cite{RL_survey,SSRL_survey}.

\begin{figure*}[t]
  \centering
  \includegraphics[width=0.99\linewidth]{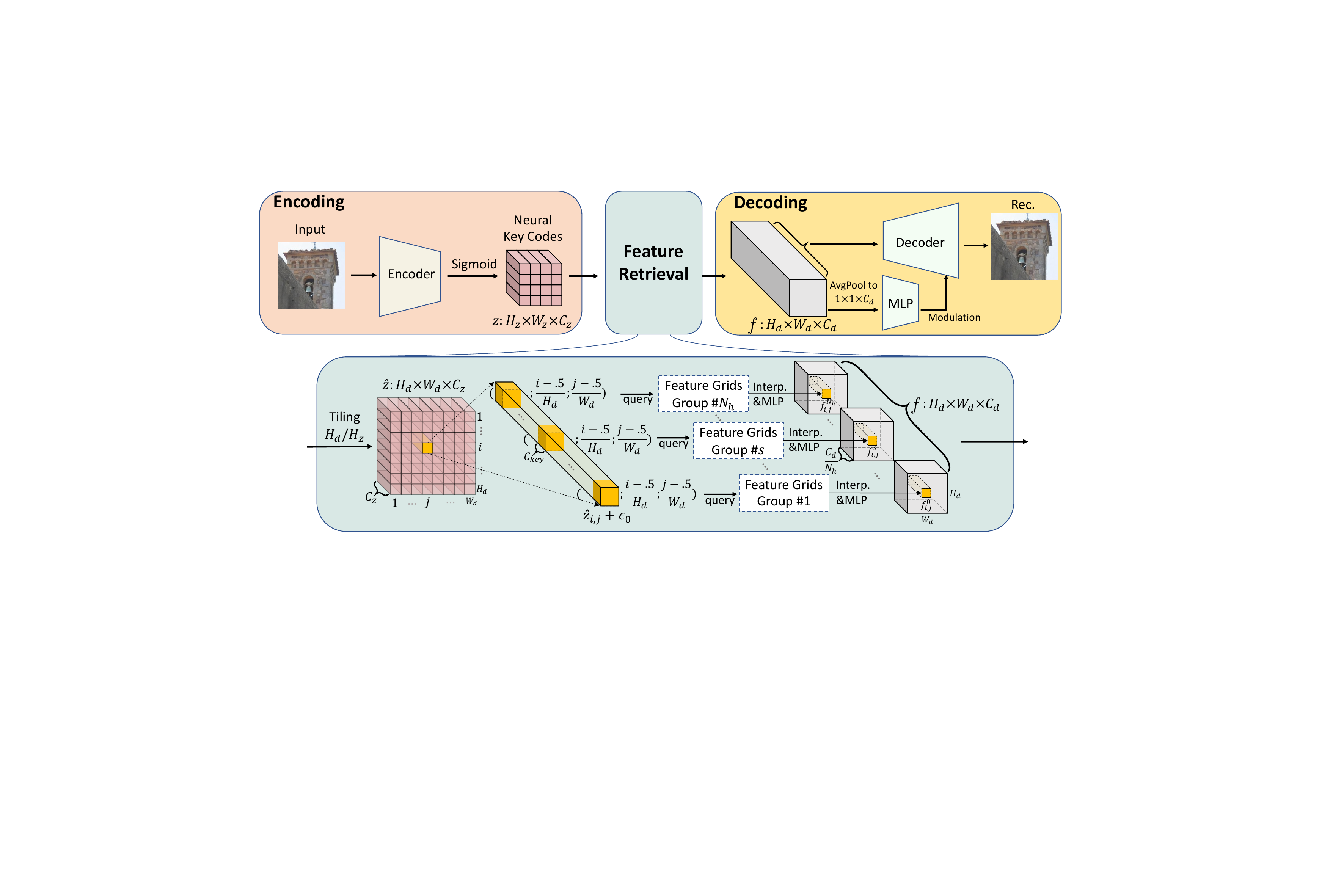}
  \caption{Overall pipeline of our method in three parts: Encoding, Feature Retrieval and Decoding.}
  \label{fig:overall_pipeline}
\end{figure*}

One notable trend from recent advances in feature representations that combines compactness, rapid training, and quality reconstruction is to map data instances 
to coordinates or coefficients whose bases are stored in auxiliary data structures, such as TensorRF~\cite{chen2022tensorf}, Instant-NGP~\cite{muller2022instant}, and Dictionary Fields~\cite{chen2023dictionary}. 
However, these mappings have so far been designed for a \emph{single instance}. How to extend such single-instance techniques to represent a dataset and how to combine it with the state-of-the-art generative methods such as diffusion models are non-trivial and worth exploring.

In this paper, we aim to develop a feature representation for an image {\em collection\/} which retains the advantages 
of the single-instance techniques, while enabling {\em generative\/} modeling over the key codes 
using the state-of-the-art generative method, the diffusion model~\cite{ho2020denoising,song2020score}; see ~\autoref{fig:best_showoff}.
To this end, we design our feature representation to be {\em bi-level\/}, consisting of a {\em key code\/} space on top of a space of
{\em multi-resolution feature grids\/}. The two spaces should be coherently connected, 
allow end-to-end training, and suitably support diffusion-based image generation over the key codes, without compromising compactness or efficient 
access. In addition, the feature grid space must accommodate diversity of the image collection with a sufficiently high dimension, with which we could face a "curse of dimensionality" if only one set of data structures is employed.


To fulfill all the criteria above and address the ensuing challenges, we train a novel autoencoder, coined \abbrv\ for Bi-level feature Representation of Image collectionS, to represent an image collection, as illustrated in~\autoref{fig:concept_of_method} and~\autoref{fig:overall_pipeline}.
Our key idea is that, via an encoder, we first map each image, in a {\em continuous\/} fashion, into a high-dimensional index block which we refer to as a key code. 
Then, we use the key codes to retrieve features from, not one, but {\em groups\/} of auxiliary feature grids.
With such a grouped data organization and the autoencoding training, \emph{compactness} is inherently established within 
the feature space, alleviating the curse of dimensionality.

Moreover, unlike VQVAE~\cite{VQVAE}, which is discrete, the continuous nature of the key codes enables concurrent training of our autoencoder, retrieval indices, and feature grids.
%
Indeed, our key codes and feature grids are jointly trained continuously with well-defined gradient flows, leading to high usage rates of the feature grids (see Section \ref{sec:properties}) and improved generative modeling compared to VQ.
%
In addition, the key codes are enlarged through a \emph{tiling} mechanism and concatenated with spatial coordinates to retrieve larger and spatial aware features, helping to reduce the size of the decoder ($\approx$50\% fewer GFlops compared to VQGAN~\cite{esser2021imagebart}) while maintaining reconstruction quality.

Furthermore, We demonstrate the applicability of our framework to generative modeling by applying the diffusion model~\cite{ho2020denoising,song2020score} on the encoded key codes learned from an image collection.
Our key codes naturally fall into the range of $[0,1]$ with bounded variance since they are the continuous indices used for feature retrieval. This property obviates the need of extra scaling operations in KL-regularized continuous latent codes~\cite{rombach2022high,chou2022diffusionsdf}. Last but not least, to address the issue of Gaussian error accumulation during the diffusion denoising process~\cite{li2023alleviating,zhang2023lookahead} (i.e. the exposure bias), we propose a relaxed precision design by injecting certain amount of Gaussian noise based on the maximum resolution of feature grids into key codes when training our autoencoder. As shown in Section~\ref{sec:properties}, this can notably benefit the diffusion training process on key codes. In the end, our generative model achieves state-of-the-art performance for image synthesis on the LSUN-Church ($\approx$29\% lower CLIP-FID, $\approx$19\% higher precision score than LDM~\cite{rombach2022high} and $\approx$44\% lower CLIP-FID than Projected GAN~\cite{sauer2021projected}) and FFHQ datasets. 




%% file: 02_RW.tex
\input{tables/symbol_table.tex}
\section{Related Work}
\label{sec:related}

At the core of \abbrv, we employ feature grids with continuous key codes for image representation and generation. Hence in this section, we focus on discussing the most relevant topics, namely, VQ latent representations, image encoding methods, as well as modern image generators.

\subsection{Vector Quantized Latent Representation}
\label{sec:vq_methods}
VQ-VAE~\cite{van2017neural} proposed to represent the latent codes of data as combinations of discrete codebook entries.
VQ-GAN~\cite{esser2021taming} proposed to add a discriminator and perceptual loss to evaluate the reconstructed images, and employed Transformers \cite{vaswani2017attention} to model the relationship of codebook entries.
RQ-VAE~\cite{lee2022autoregressive} used a residual quantizer to represent an image as a stacked map of discrete codes, resulting in a short sequence of latent codes compared to VQ-GAN. MoVQ~\cite{zheng2022movq} proposed to incorporate the spatial information into the quantized vectors so that similar patches within an image do not share the same index.
Vanilla VQ methods (VQ-VAE and VQ-GAN) often face the issue of low codebook usage because of \circled{1} poor initialization of the codebook, which means only a few codebook entries will be used in the very beginning and \circled{2} the training procedure does not guarantee the full coverage of the codebook entries. A codebook entry $e$ only gets updates from the features that are assigned to it. So dead codes cannot be updated. VIT-VQGAN~\cite{yu2021vector} aims to enhance codebook utilization by reducing its dimensionality and applying $\mathcal{L}_{2}$ normalization to the encoded latent vectors.
 
Our method differs from VQ methods as it performs interpolations on the entries in feature grids. This makes the mapping from key codes to feature vectors continuous. Furthermore, we utilize an ordinary stochastic gradient descent algorithm to optimize the entry values in feature grids. Compared to the clustering-based optimization strategy in VQ methods that suffer from usage rate, our method fully utilizes the entries (see Section~\ref{sec:properties} and Section~\ref{sec:high-hit-ratio} of the supplementary), thus offering a more powerful representation.

\subsection{Image Synthesis via Latent Diffusion}
Recently, diffusion models~\cite{sohl2015deep,ho2020denoising,kingma2021variational,dhariwal2021diffusion} have demonstrated their superiority over  VAE~\cite{kingma2013auto}, GANs~\cite{goodfellow2020generative}, Flow-based models~\cite{kingma2018glow} in terms of sample quality, ease of training, and scalability.
Latent Diffusion Models~\cite{rombach2022high} (LDM) have been proposed to perform diffusion process on the latent codes of images instead of raw pixels. 
In a vanilla LDM, diffusion is usually performed on the indexings of VQ codebook entries. 
However, as mentioned above, the VQ codebooks usually suffer from low usage rates. In contrast, the key codes from our representation admit much higher utilization of the feature grid entries with improved expressivity, bounded variance and a relaxed precision that can all benefit the following diffusion process.

\subsection{Image Encoding by Neural Representation}

Single instance fitting method has been studied on how to get a compact and accurate representation on that instance. Works~\cite{chen2019learning,park2019deepsdf} use Fourier positional encodings as input with an MLP to represent 3D SDF field, which can also be applied to an image. SIREN~\cite{sitzmann2020implicit} utilizes MLP with periodic functions to represent an image or a 3D shape. Instant NGP~\cite{muller2022instant}, employs multi-resolution hash encoding to accelerate the prediction of graphics primitives such as color, volume density, and signed distance field.

%% file: tables/symbol_table.tex
\begin{table*}[t]
    \small
    \centering
    \resizebox{\textwidth}{!}{
    \begin{tabular}{clcl}
        \thickhline 
        \addlinespace[2pt]
        Symbol & Description & Symbol & Description \\
        \addlinespace[2pt]
        \hline
        \addlinespace[2pt]
        $z$ &  key code encoded from input image & $F_r$ &  spatial feature retrieval function \\
        $\hat{z}$ &  interleavingly tiled key code & $F_{inp}$ &  linear interpolation function \\
        $\hat{z}_{i,j}$ &  tiled key code at spatial location $(i,j)$ & $F_{mlp}$ & the small MLP network in each group of feature grids \\
        $H_z,W_z$ &  spatial size of the key code & $N_h$ &  number of groups of feature grids \\
        $H_d,W_d$ &  spatial size of the tiled key code & $N_r$ & the resolution number of feature grids \\
        $C_z$ &  channel size of the key code & $C_{key}$ &  length of a unit key code \\
        $C_d$ &  channel size of the feature block to decoder & $f_k^v$ &  feature vectors that encompass $q_{i,j}^s$ in feature grids \\
        $\hat{z}^s_{i,j}$&  $s$-th slice of tiled key code $\hat{z}_{i,j}$ & $c^v_k$ &  coordinates of $f_k^v$ \\
        $q_{i,j}^s$ & $\hat{z}^s_{i,j}$ combined with spatial coordinates & $f_{i,j}^s$ &  retrieved feature vector for every $q_{i,j}^s$ \\
        \addlinespace[2pt]
        \hline
    \end{tabular}}
    \vspace{4pt}
    \caption{Mathematical symbols utilized in our paper along with their descriptions.}
    \label{tab:symbol_table}
\end{table*}

%% file: 03_method.tex
\section{Method}
\label{sec:method}

We propose a novel and general method to represent a dataset by continuous key codes and groups of feature grids within an autoencoder framework. The feature space can be compactly and structurally organized in the multi-resolution feature grids. Any image can be encoded by referring to its key code and then performing interpolations on its relevant feature vectors in feature grids.

Our method is comprised of three main components: \textit{\textbf{Encoding}}, \textit{\textbf{Feature Retrieval}}, and \textit{\textbf{Decoding}} (~\autoref{fig:concept_of_method}). The overall pipeline is shown in ~\autoref{fig:overall_pipeline}. In the following, we discuss these components in detail and explain their respective functions. We have also provided our mathematical symbols and their descriptions in \autoref{tab:symbol_table}.


\subsection{Encoding}
\label{sec:encoding}
Given an image $I\in \mathbb{R}^{H \times W \times C}$ with dimension $H\times W$ and channel width $C$, an encoder network is utilized to transform it into a 
{\em key code\/} $z\in [0, 1]^{H_z \times W_z \times C_z}$ in a {\em continuous\/} space, where $C_z = N_h \times C_{key}$ with $N_h$ being the number of groups of 
feature grids and $C_{key}$ the length of a key unit, each corresponding to one of the feature grids. 
Our encoder follows the design of VQ-GAN~\cite{esser2021taming}, except that a sigmoid function is appended to normalize $z$ into $[0, 1]$.
This continuous representation $z$ serves as the key to access the multi-resolution feature grids, which provide a grid-based representation of image features over the entire image collection.

\subsection{Feature Retrieval}
\label{sec:Hash_Retrieval}

The varying levels of detail in a multi-resolution feature grid are controlled by image resolution.
Our feature retrieval allows the continuous key codes to be used to retrieve 
features stored in our multi-resolution feature grids for an entire dataset.
The retrieved entries are linearly interpolated to obtain the image features at the given key code.

Before retrieval from the feature grids, we first tile our continuous code $z$ interleavingly to form a tensor $\hat{z}$ of size $H_d \times W_d \times C_z$, where $H_d$ ($\ge H_z$) and $W_d$ ($\ge W_z$) are the initial spatial resolution for the decoder (see~\autoref{fig:overall_pipeline} (Bottom)). The tiling is used to enlarge the spatial size of features that will be sent to decoder. A larger spatial size will result in a smaller and more efficient decoder than previous methods~\cite{esser2021taming,zheng2022movq,teschner2003optimized}.
For each spatial grid $\hat{z}_{i,j}$ in $\hat{z}$, we split its $C_z$ channels into $N_h$ slices of codes and each code is in the shape of $1 \times 1 \times C_{key}$, in which we can utilize multiple groups of feature grids to increase the model capacity and alleviate the computational cost of feature interpolation.
A small noise $\epsilon_o \sim \mathcal{N}(0, \frac{1}{(4r_{max})^2})$ is injected in each code slice $\hat{z}_{i,j}^{s}$ ($1 \le s \le N_h$), where $r_{max}$ denotes the maximum resolution of feature grids. Perturbed $\hat{z}_{i,j}^{s}$ is then concatenated with its normalized spatial coordinates $(x = \frac{i - 0.5}{H_{d}}, y = \frac{i - 0.5} {W_{d}})$ to form a combined key code: 
\begin{equation}
q_{i,j}^{s} = \left[\hat{z}_{i,j}^{s} + \epsilon_o;x;y\right],
\end{equation}
where $s$ is the index of slices and $q_{i,j}^{s} \in [0, 1]^{C_{key} + 2}$. 
The minor random perturbation is to construct a relaxed data precision of $\hat{z}_{i,j}^{s}$, which will ease the diffusion training in the second stage (explanations are given in \autoref{sec:reasons_data_precision}).
Incorporating $(x, y)$ into key codes aims to retrieve spatial-aware features from multi-resolution feature grids.
The retrieved features $f_{i,j}^{s}$ are then placed into the corresponding location of the feature block $f\in \mathbb{R}^{H_d \times W_d \times C_d}$, which serves as the input features to the decoder network (see~\autoref{fig:overall_pipeline}).
Mathematically, $f_{i,j}^{s}$ is obtained as follows:
\begin{equation}
f_{i,j}^{s} = F_{mlp}(F_{inp}(F_r(q_{i,j}^{s}))),
\end{equation}
where, $F_r$ denotes the retrieval function returning the grid nodes that encompass key code $q_{i,j}^{s}$ in feature grids. For more details on $F_r$, please see section~\ref{sec:detailed_AE} of the supplementary. The result of $F_r$ is a set of tuples $(f_{k}^{v}, c_{k}^{v})$ where $f^{v} \in \mathbb{R}^{N_r \times C_v}$ contains feature vectors and $c^{v} \in [0, 1]^{C_{key} + 2} $ is their associated coordinates. $N_r$ is the number of resolutions in each group of feature grid and $C_v$ is the dimension of feature vectors in feature grids. $F_r$ is defined as:
\begin{equation}
F_r(q_{i,j}^{s})=\{(f_{k}^{v}, c_{k}^{v}) | 1 \le k \le M \} .
\end{equation}
Here, $M=2^{C_{key} + 2}$. For example, if $C_{key} =1$, $c_{k}^{v}$ will represent the coordinates of a cube and $f_{k}^{v}$ will be the eight feature vectors associated with each of the cube's nodes.

$F_{inp}$ indicates a linear interpolation among $f_{k}^{v}$s returning a feature vector of size $N_r \times C_v$ by:
\begin{equation}
\sum_{k=1}^{M} F_d(c_{k}^{v}, q_{i,j}^{s}) \cdot f_{k}^{v},
\end{equation}
where $F_d$ is a distance function from $c_{k}^{v}$ to $q_{i,j}^{s}$, determining the weight of feature vector $f_{k}^{v}$. When $C_{key} =1$, $F_{inp}$ is a tri-linear interpolation on the cube's nodes.
Finally, $F_{mlp}$ maps the dimension of interpolated feature vector from $N_r \times C_v$ into $C_d / N_h$ using an MLP.

\begin{figure*}[h]
  \centering
    \begin{subfigure}[b]{0.23\textwidth}
    \centering
    \includegraphics[width=\textwidth]{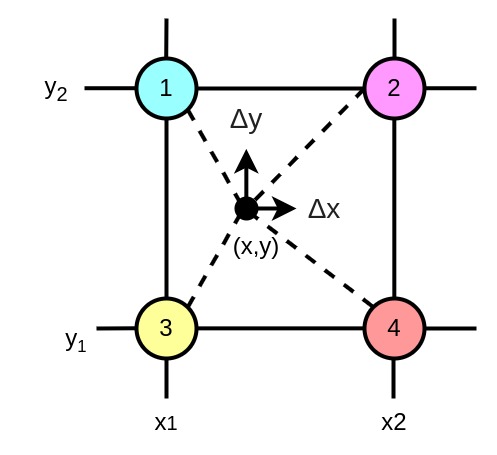}
    \newsubcap{The continuous updating mechanism in our method by a linear interpolation of the values of nearby grid nodes.}
    \label{fig:continuous_interp}
  \end{subfigure}
  \hspace{1pt}
  \begin{subfigure}[b]{0.35\textwidth }
    \centering
    \includegraphics[width=\textwidth]{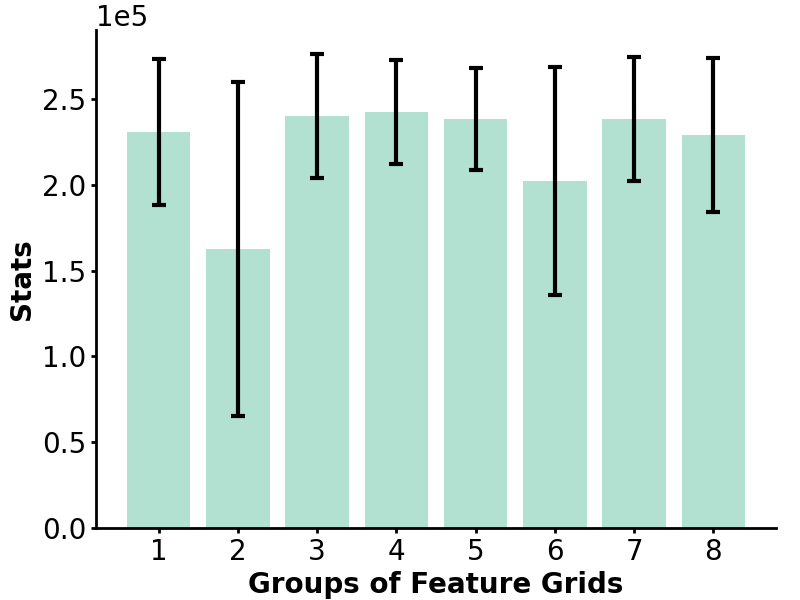}
    \newsubcap{The hitting statistics of feature grid entries are shown at the maximum resolution across 8 groups of feature grids ($N_h=8, C_z=8, C_{key}=1$) on the entire training dataset.}
    \label{fig:our_hitting_ratio}
  \end{subfigure}
  \hspace{1pt}
  \begin{subfigure}[b]{0.36\textwidth}
    \centering
    \raisebox{8mm}{
    \includegraphics[width=\textwidth]{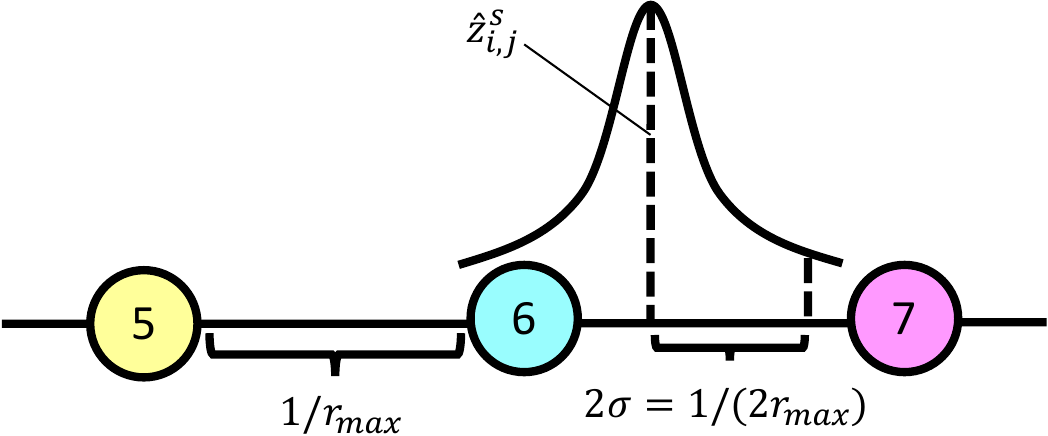}}
    
    \newsubcap{Relaxed precision mechanism of our method illustrated in 1D grid. $r_{max}$ is the maximum resolution in multi-resolution feature grids and the colorful circles are grid's nodes.}
    \label{fig:controlled_procesion}
  \end{subfigure}
\end{figure*}

\subsection{Decoding}
\label{sec:Decoding}
The architecture of our decoder follows StyleGAN~\cite{karras2020analyzing}. The decoder is modulated by a style code $\in \mathbb{R}^{1 \times 1 \times C_d}$, which is calculated from $f$ by average pooling and MLPs.

The entire autoencoding process is trained using ($\mathcal{L}_{2}$) loss , perceptual loss($\mathcal{L}_{percep}$)~\cite{johnson2016perceptual}, Wasserstein GAN loss ($\mathcal{L}_{WGAN}$)~\cite{arjovsky2017wasserstein} and discriminator's gradient penalty ($\mathcal{L}_{gp}$)~\cite{gulrajani2017improved}. The total loss is: 
\begin{equation}
    \mathcal{L}_{total} = \mathcal{L}_{2} + \mathcal{L}_{percep} + \mathcal{L}_{WGAN} + \mathcal{L}_{gp},
\end{equation}
to jointly optimize the encoder, decoder and feature grids.

\subsection{Properties}
\label{sec:properties}

Here, we discuss four key properties of our method: {\circled{1}~\emph{auxiliary data structure}, \circled{2}~\emph{continuity}, that are beneficial for reconstruction and   \circled{3}~\emph{relaxed data precision}, and \circled{4}~\emph{strictly bounded variance} that improve generation. We will explain these four properties in the following.

\noindent
\textbf{Auxiliary Data Structure.} Rather than directly passing the encoded features to the decoder for reconstruction, we encode the keys used to retrieve features from an auxiliary data structure, i.e., the multi-resolution feature grids. The multi-resolution feature grids provide powerful feature composition-ability together with spatial coordinate codes while other continuous latent codes~\cite{rombach2022high,chou2022diffusionsdf} do not. 

\noindent
\textbf{Continuity.} 
VQ methods~\cite{van2017neural,esser2021taming} involve using a metric to determine the closest code in a codebook to the encoded feature. The identified code's gradient is directly applied to the encoded feature during backpropagation. 
Unlike VQ, our key code and feature grids are jointly trained continuously with well-defined gradient flows. As illustrated in 2D grids of \autoref{fig:continuous_interp}, the retrieved values at input key code $(x,y)$ are linearly interpolated by its four encompassing grid nodes and moved in the direction $\Delta x = [(y_2 - y)(Q_4 - Q_3) + (y-y_1)(Q_2 - Q_1)]\cdot \mathrm{grad}$, where $Q_i$ is the value of nodes (a similar derivation can be found for $y$ direction). 

On the other hand, studies~\cite{yu2021vector,zeghidour2021soundstream} have shown that the original VQ method tends to have a low codebook usage ratio. The discrete nature of VQ requires performing careful tricks including a special initialization scheme of the codebook, an alternative distance metric~\cite{yu2021vector} and top-k and top-p (nucleus) sampling heuristics~\cite{esser2021taming}.
In contrast, as shown in \autoref{fig:our_hitting_ratio}, all entries in our feature grids are hit and the hitting variance is very small in most groups of feature grids. More analysis on the high usage ratio of feature grids is available in section~\ref{sec:high-hit-ratio} of the supplementary material. Moreover, the continuity here also allows us to introduce relaxed data precision with a controlled scale of noise perturbation, which will be discussed in the following part.

\noindent
\textbf{Relaxed Data Precision.} 
\label{sec:reasons_data_precision}
To use our method for image generation, we apply diffusion model on the continuous key codes. Due to the practical setting (finite diffusion steps) of the diffusion model and the limited capability of the diffusion network, it has been reported~\cite{li2023alleviating,zhang2023lookahead} that there exist accumulated errors that conform to a normal distribution between the predicted $\tilde{x}_t$ and the ground truth $x_t$ (the exposure bias). In other words, we are unable to approach data in arbitrary precision on the current diffusion setting. To alleviate this issue, we inject a noise $\epsilon_o \sim \mathcal{N}(0, \frac{1}{(4r_{max})^2})$ to key codes $\hat{z}_{i,j}^s$ as illustrated in~\autoref{fig:controlled_procesion} for a simplified 1D grid. After training, the autoencoder learns to produce reasonable reconstruction results from the perturbed key codes as long as they are within the interval of $\hat{z}_{i,j}^s \pm \frac{1}{2r_{max}}$. We refer to this design choice as a {\em relaxed data precision\/}. With default $r_{max}=64$, our relaxed precision is even lower than ordinary 8bit images. We have shown in \autoref{sec:results} within our bi-level representation framework, the relaxed data precision trick is robust for reconstruction, makes it easier for diffusion model and can generate better results. More analysis is provided in section~\ref{sec:relaxed-precision} of the supplementary material.

\noindent
\textbf{Strictly Bounded Code Variance.}
As shown in prior latent diffusion models \cite{rombach2022high,chou2022diffusionsdf}, KL regularization is utilized to prevent them from growing to codes of arbitrary high variance. However, KL regularization alone cannot ensure a proper upper bound of latent code variance. Excessive weights on KL regularization may degrade the reconstruction. Normalizing each latent code separately, without taking into account the data distribution, will also be detrimental as it disrupts learning the distribution. To address these issues, an element-wise rescaling approach is used on KL-regularized latent codes when training diffusion models, where the rescaling factor is computed from the very first mini-batch of training data~\cite{rombach2022high}. In comparison, our continuous latent codes are keys of feature grids and they are therefore naturally in the range $[0, 1]$, and have a strictly bounded variance.

%% file: 04_results.tex
\section{Results and Evaluation}
\label{sec:results}


\textbf{Implementation Details and Datasets.}
Our experiments are conducted on LSUN-Church~\cite{yu2015lsun} and FFHQ~\cite{karras2019style} datasets. Images in both datasets are center cropped and resized to  $256 \times 256$ as in previous works~\cite{sauer2021projected,karras2020analyzing,karras2021alias,rombach2022high}.
Our encoder's architecture is borrowed from the one of VQGAN~\cite{esser2021taming}, while our decoder and discriminator are based on the architecture of StyleGAN2~\cite{karras2020analyzing}.
 We use the unit indexing dimension $C_{key}=1$ and the spatial size of $f$ in the decoder as $H_d=W_d=64, C_d=512$ across all the experiments. There are $N_r=16$ resolutions in each group of feature grids starting from 4 to 32 or 64 ($r_{max}$). Each feature grid has maximum $2^{18}$ entries where each entry contains four float numbers ($C_v = 4$).

\input{tables/reconstruction_result_table}
\input{tables/computational_cost}
\input{tables/lsun_church_quantitative_results}
\input{tables/ffhq_quantitative_results}

For the diffusion models, we utilize the typical UNet used in networks such as Imagen~\cite{saharia2022photorealistic} with unit channel size 256. The UNet is predicting noise \cite{ho2020denoising,song2020score} with either cosine noise scheduler~\cite{nichol2021improved} and min-SNR weighting strategy~\cite{hang2023efficient} or a variant of cosine noise scheduler ~\cite{chen2023importance}. More implementation details of autoencoding and diffusion training are provided in section~\ref{sec:more_implementation_defails} of the supplementary material.

\begin{figure*}
    \centering
    \includegraphics[width=0.98\textwidth]{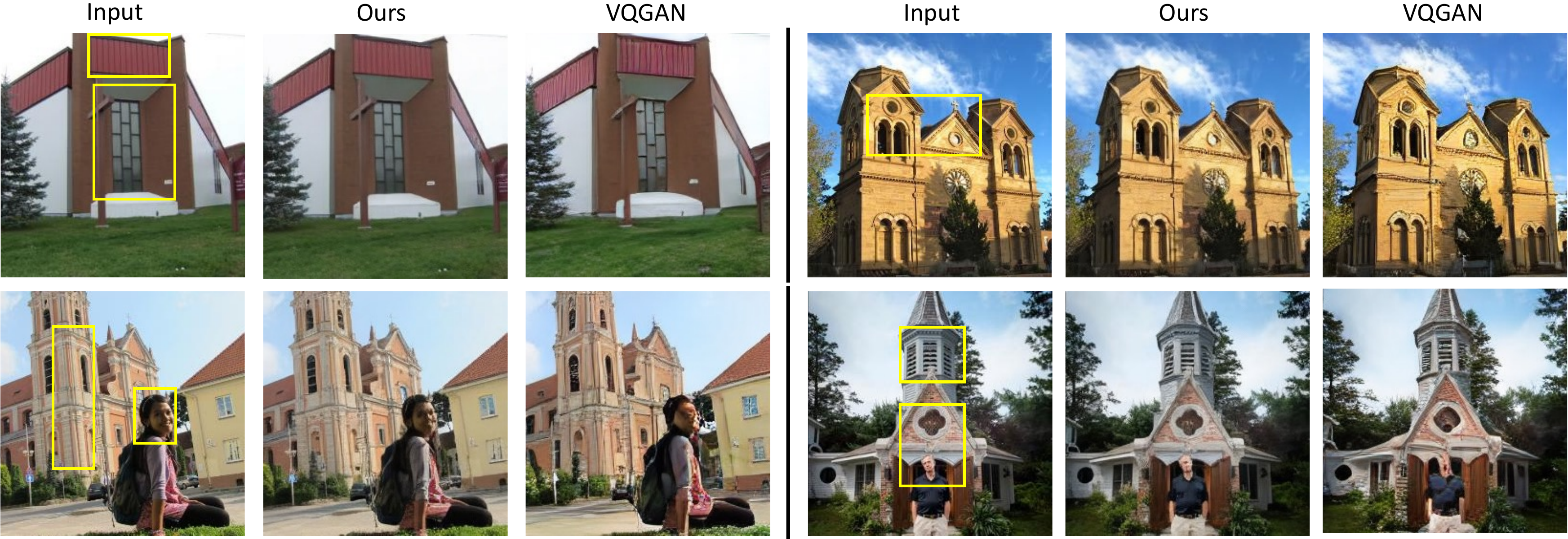}
    \caption{Comparisons to VQGAN~\cite{esser2021taming} demonstrate that our method consistently produces superior reconstruction quality. For optimal viewing and contrast, zoom in on highlighted regions corresponding to the \textcolor{yellow}{yellow} squares.}
    \label{fig:reconstruction_comparison}
\end{figure*}

\noindent
\textbf{Comparisons.}
For image reconstruction, we compare with SOTA methods including VQ-GAN~\cite{esser2021taming},  RQ-VAE \cite{lee2022autoregressive}.
For image synthesis, we compare with Latent Diffusion~\cite{rombach2022high}.
Our competitive performance becomes evident when examining the reconstruction metrics (LPIPS~\cite{zhang2018unreasonable}, SSIM~\cite{wang2004image}, and PSNR) as shown in \autoref{tab:recon_quantative_results}. Notably, our approach achieves these commendable results while utilizing a much smaller and more efficient decoder, as shown in \autoref{tab:model_params_and_computation}. The qualitative comparison of reconstruction results with VQGAN~\cite{esser2021taming} is shown in \autoref{fig:reconstruction_comparison} indicating that our results have less noise and better preserved details.

\begin{figure*}
\centering
\includegraphics[width=1.0\textwidth]{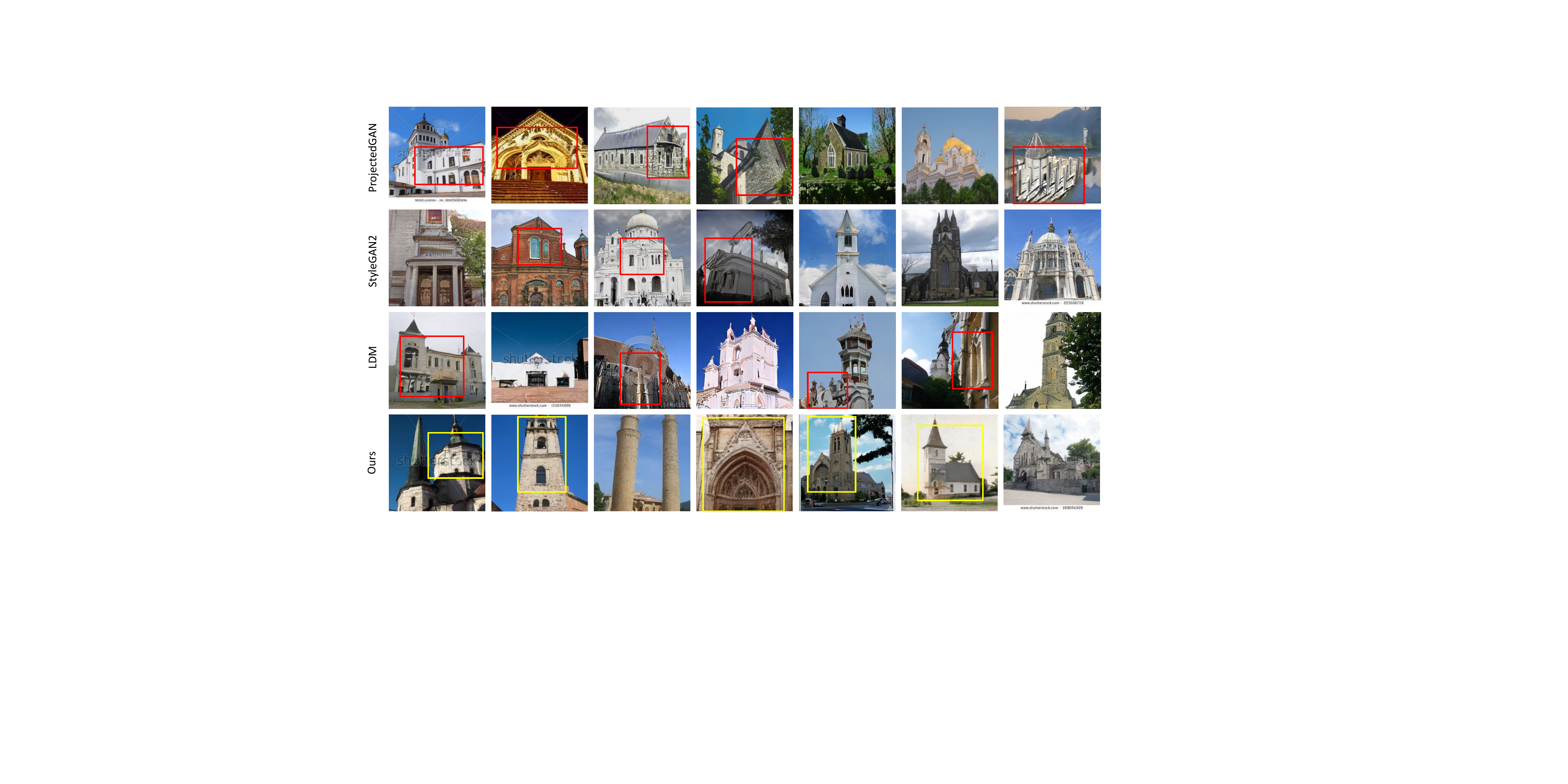}
\caption{Image generation by Projected GAN~\cite{sauer2021projected}, StyleGAN2~\cite{karras2020analyzing}, LDM~\cite{rombach2022high}, and our method on 256$\times$256 LSUN-Church dataset~\cite{yu2015lsun}.
All results shown were selected {\em randomly\/}. 
Our method clearly outperforms the others in terms of structural coherence and regularity, e.g., less distortion, symmetries (see \textcolor{yellow}{yellow} squares). \textcolor{red}{Red} squares highlight significant distortion or structural issues. More results are shown in the section~\ref{sec:more_LSUN_church_results} of supplementary material.}
\label{fig:generation_comparison}
\end{figure*}

For generation, we apply diffusion on the key codes of size $64\times64\times4$ on LSUN-Church dataset~\cite{yu2015lsun} and FFHQ~\cite{karras2019style}.
Our method is capable of generating high-quality images. The lines of the church building in our results are more straight than those in LDM~\cite{rombach2022high} and our windows are more regularly patterned than StyleGAN2~\cite{karras2020analyzing} and Projected GAN~\cite{sauer2021projected} (~\autoref{fig:generation_comparison}). For more visual results, please see sections~\ref{sec:more_FFHQ_results} and~\ref{sec:more_LSUN_church_results} of the supplementary material. 

\begin{figure}
    \centering
    \includegraphics[width=0.45\textwidth]{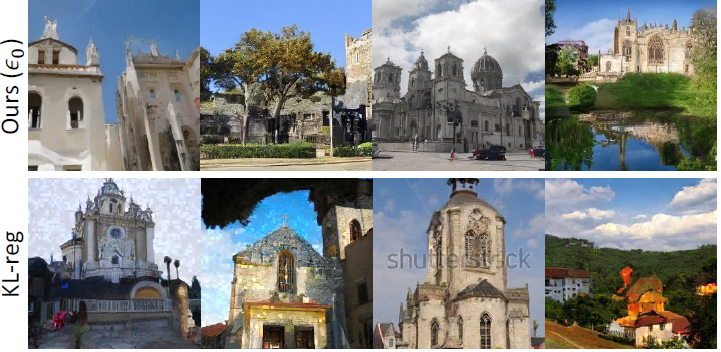}
    \caption{The generated results of diffusion model applied on KL-regularized key codes versus ours when using cosine noise scheduler with min-snr weighting strategy. KL-reg results show irregular spot artifacts.}
    \label{fig:KL_results}
\end{figure}

We also measure the generation results quantitatively in the metrics of FID~\cite{heusel2017gans}, CLIP-FID~\cite{kynkaanniemi2022role}, Inception Score (IS)~\cite{salimans2016improved}, and Precision-Recall~\cite{kynkaanniemi2019improved}. F1-score is the harmonic mean of precision and recall scores. Metrics are evaluated on 50k generated images and more details of the evaluation can be found in section~\ref{sec:evaluation_details} of the supplementary.
We show quantitative results on the generations in LSUN-Church and FFHQ datasets in \autoref{tab:ffhq_quantitative_results} and \autoref{tab:lsun_church_quantitative_results}. In \autoref{tab:lsun_church_quantitative_results}, we compare four configurations, namely combinations of two noise scheduling strategies (\textbf{a.} cosine noise scheduler~\cite{nichol2021improved} and min-snr~\cite{hang2023efficient}; \textbf{b.} a variant of cosine noise scheduling~\cite{chen2023importance}) with two regularization methods on key codes (\textbf{c.} KL-regularization on key codes; \textbf{d.} relaxed data precision with noise perturbation $\epsilon_0$). The \textbf{b} + \textbf{d} configuration is the default one. As we can see from \autoref{tab:lsun_church_quantitative_results}, our default setting (Config b + d) achieves \textbf{$\approx$29\% lower} CLIP-FID score and \textbf{$\approx$19\% higher} precision score than LDM as well as \textbf{$\approx$44\% lower} CLIP-FID than Projected GAN~\cite{sauer2021projected} while maintaining competitative on all other metrics. When applying diffusion model on KL-regularized key codes with cosine noise scheduler~\cite{nichol2021improved} and min-snr weighting strategy~\cite{hang2023efficient} (Config a + c), the generated results show irregular spot artifacts in images shown in \autoref{fig:KL_results} and gets much worse quantitative results \autoref{tab:lsun_church_quantitative_results}. We owe this phenomenon to that KL-regularized key codes cannot ensure proper data precision while our method can pose controlled relaxed data precision that makes diffusion model training easier. More analysis on our relaxed data precision method is given in section~\ref{sec:relaxed-precision} of the supplementary material. 

\input{tables/reconstruction_with_noise_table}


\begin{figure*}[h]
\centering
\includegraphics[width=1.0\textwidth]{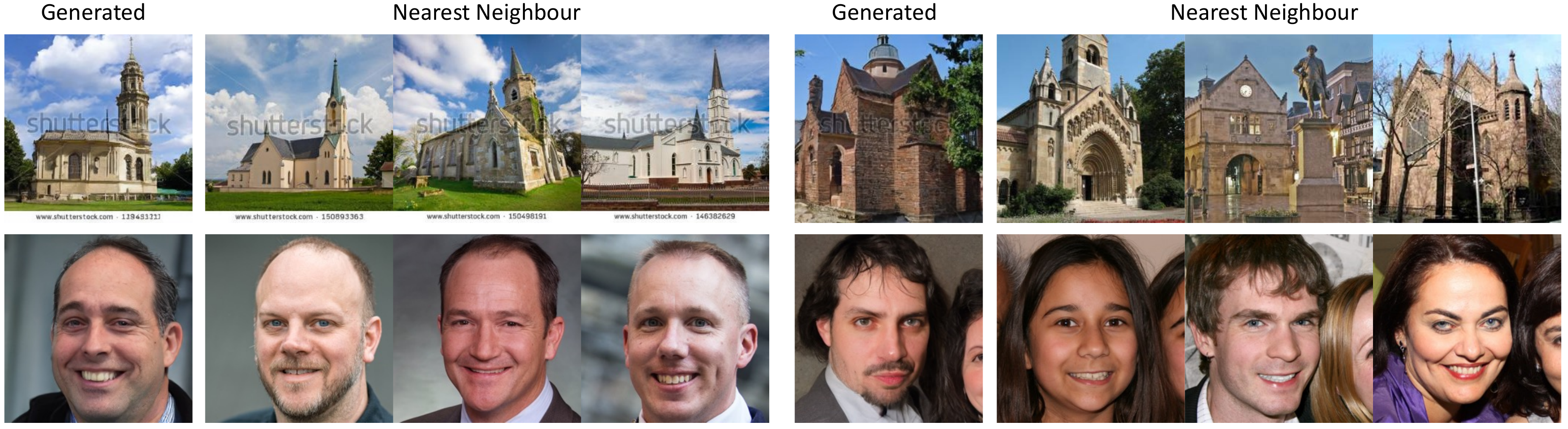}
\caption{Nearest (by LPIPS~\cite{zhang2018unreasonable}) neighbour images of generated samples from the diffusion training dataset.}
\label{fig:nearest_neighbor_search}
\end{figure*}


\noindent
\textbf{Model Size and Tradeoffs.}
The influence of code sizes on reconstruction quality and model size is shown in \autoref{tab:recon_quantative_results} and \autoref{tab:model_params_and_computation}. 
Our largest model (code size $16\times16\times16$) has only 10M more parameters ($\approx24\%$ increase) than those of the corresponding VQGAN and RQ-VAE models. However, our model outperforms VQGAN and RQ-VAE across all metrics, showing 55\% and 41\% improvements in LPIPS, respectively. In terms of computational costs, our method requires $\approx50\%$ fewer GFlops due to its smaller decoder and the efficiency gained from utilizing the feature grids. 

\noindent
\textbf{Memory and Computational Overhead}
The memory overhead by introducing our bi-level representation is negligible. The memory usage of the feature retrieval module during training (including backward) is 91.20$\pm$2.8MB for batch size 1 and 126.05$\pm$1.7MB for batch size 2. As for the computational overhead, the computations inside the feature retrieval is only 0.27 GFLOPs vs total 63.40 GFLOPs. More detailed analysis of computational and memory overhead can be found in sections~\ref{sec:computational_overhead},\ref{sec:memory_overhead} of the supplementary material.

\noindent
\textbf{Ablation Study on $\epsilon_0$.}
\label{minisec:ablation_study_on_epsilon_0}
To demonstrate the impact of the noise perturbation $\epsilon_0$ (relaxed precision), we compared the reconstruction performance with and without noise (\autoref{tab:noise_perturb_recon}) showing that the noise perturbation not only does not degrade the reconstruction on the LSUN-Church dataset~\cite{yu2015lsun} but leads to slight improvements.
These findings provide empirical evidence on the robustness of our method for the reconstruction task. For more parameter and ablation studies, please refer to sections~\ref{sec:parameter_studies},\ref{sec:compatibility_with_other_decoders},\ref{sec:compatibility_with_hash_tables} of the supplementary material.


\noindent
\textbf{Nearest Neighbour Search.}
We also compare our generated images with the nearest neighbour images obtained from the diffusion training dataset using the LPIPS distance~\cite{zhang2018unreasonable}. \autoref{fig:nearest_neighbor_search} demonstrates that our method generates distinct samples rather than retrieving from the training set.

%% file: tables/reconstruction_result_table.tex
\begin{table*}[t]
\resizebox{1.0\textwidth}{!}{
\begin{tabular}{cccccccccc}
     \thickhline
     \addlinespace[4pt] 
     \multicolumn{5}{c}{FFHQ 256$\times$256} & \multicolumn{5}{c}{LSUN-Church 256$\times$256} \\
     \addlinespace[4pt] 
     \hline 
     \addlinespace[4pt] 
     Method                                 & Code Size &  LPIPS $\downarrow$ & SSIM $\uparrow$ & PSNR $\uparrow$ &
     Method                                 & Code Size &  LPIPS $\downarrow$ & SSIM $\uparrow$ & PSNR $\uparrow$ \\ 
     \addlinespace[4pt]
     \hline
     \addlinespace[4pt]
     VQGAN \cite{esser2021taming}           & 16 $\times$ 16 & 0.1175 & 0.6641 & 22.24 &
     VQGAN \cite{esser2021taming}           & 32 $\times$ 32 & 0.14$\pm$0.07 & 0.53$\pm$0.23 & 19.81$\pm$4.92  \\
     RQ-VAE \cite{lee2022autoregressive}    & $16 \times 16 \times4$ & 0.0895 & 0.7602 & 24.53 &
     Ours                                   & 32 $\times$ 32 $\times$ 4 & 0.12$\pm$0.07 & 0.67$\pm$0.20 & 22.32$\pm$5.93\\
     Ours                                   & $16 \times 16 \times8$ & 0.0885 & 0.7178 & 25.72 &
     Ours                                   & 32 $\times$ 32 $\times$ 8 & 0.07$\pm$0.05 & 0.76$\pm$0.17 & 24.23$\pm$6.62\\
     Ours                                   & $16 \times 16 \times16$&\textbf{0.0522} & \textbf{0.7651} & \textbf{26.99}& 
     Ours                                   & $64\times64\times4$ & \textbf{0.04$\pm$0.03} & \textbf{0.84$\pm$0.13} & \textbf{26.34$\pm$7.14}\\
     \addlinespace[4pt]
     \thickhline
\end{tabular}
}
    \vspace{5pt}
     \caption{Reconstruction metrics on the validation splits of FFHQ~\cite{karras2019style} and LSUN-Church dataset ~\cite{yu2015lsun}.}
     \label{tab:recon_quantative_results}
\end{table*}

%% file: tables/computational_cost.tex
\begin{table*}[t]
    \resizebox{1.0\textwidth}{!}{
    \begin{tabular}{cccccccc}
     \thickhline
     \addlinespace[4pt] 
     Method                                 & VQGAN~\cite{esser2021taming}  & VQGAN~\cite{esser2021taming} 
                                            & RQ-VAE~\cite{lee2022autoregressive} & Ours 
                                            & Ours & Ours & Ours \\ \hline
     \addlinespace[4pt] 
     Code Size                              & 16$\times$16 & 32$\times$32 & $16\times16\times4$ &  $16\times 16 \times 8$ & $16\times 16 \times 16$ 
                                            & $32\times 32 \times 4$ & $64\times 64 \times 4$ \\
     Decoder \#Parameter                    & 42.5M & 13.6M & 42.5M & 21.2M$^*$ + 9.9M & 42.9M$^*$ + 9.9M  
                                            & 11.4M$^*$ + 9.9M & 18.6M$^*$ + 9.9M\\
     Deocder GFlops                         & 126.79 & 119.77 & 126.79 & 63.54$^\ddagger$ & 63.80$^\ddagger$ & 63.40$^\ddagger$ & 63.41$^\ddagger$\\
     \addlinespace[4pt]
     \thickhline
\end{tabular}}
     \vspace{5pt}
     \caption{Trainable parameters and computational load of decoders. An $^*$ indicates total number of parameters in feature grids and $^\ddagger$ refers to total computational cost of decoding {\em and\/} feature retrieval from feature grids.} 
    \label{tab:model_params_and_computation}
\end{table*}

%% file: tables/lsun_church_quantitative_results.tex
\begin{table}[!htb]
    \centering 
    \small
    \resizebox{0.5\textwidth}{!}
    {
    \begin{tabular}{ccccccc}
     \thickhline
     \addlinespace[4pt] 
     Method                                 & FID $\downarrow$ & CLIP-FID $\downarrow$  & IS $\uparrow$
                                            & Precision $\uparrow$ & Recall $\uparrow$ & F1-score $\uparrow$ \\
     \addlinespace[4pt] 
     \hline
     \addlinespace[4pt] 
     DDPM~\cite{ho2020denoising}            & 7.89 & -
                                            & - & - & - & - \\
     Projected GAN$^*$~\cite{sauer2021projected}    &       \makecell{\textbf{3.39}} &                                                9.04
                                            & \textbf{2.94} & 0.60 & \textbf{0.52} & \underline{0.561} \\
     StyleGAN2~\cite{karras2020analyzing}    & \underline{3.86} &                                                7.46 
                                            & \underline{2.76} & 0.63 & 0.43 & 0.511 \\
     LDM~\cite{rombach2022high}             & 4.02 & 7.15 
                                            & 2.72 & 0.64 & \textbf{0.52} & \textbf{0.573} \\
     Ours (Config a + c)                    & 7.84 & 8.68 
                                            & 2.55 & 0.70 & 0.36 & 0.475 \\
     Ours (Config a + d)                    & 4.66 & 8.52 
                                            & 2.56 & \textbf{0.78} & 0.41 & 0.537 \\
     Ours (Config b + c)                    & 4.50 & \underline{6.54} 
                                            & 2.56 & 0.75 & 0.44 & 0.555 \\
     Ours (Config b + d)                    & 3.97 & \textbf{5.10} 
                                            & 2.59 & \underline{0.76} & \underline{0.46} & \textbf{0.573} \\
     \addlinespace[4pt]
     \thickhline
    \end{tabular}
    
}
     \vspace{5pt}
     \caption{Quantitative results of generation on LSUN-Church dataset~\cite{yu2015lsun}. Our relaxed precision (Config d) method is adapted to two distinct noise schedulers (Config a,b) and gets record low CLIP-FID in Config b + d. $^*$ denotes that we measure the metrics of Projected GAN~\cite{sauer2021projected} using the checkpoint provided from the official Projected GAN Github repository. Underlined numbers are the second best results.}
    \label{tab:lsun_church_quantitative_results}
\end{table}

%% file: tables/ffhq_quantitative_results.tex
\begin{table}[htb]
    \centering 
    \small
    \resizebox{0.5\textwidth}{!}
    {
    \begin{tabular}{ccccccc}
     \thickhline
     \addlinespace[4pt] 
     Method                                 & FID $\downarrow$ & CLIP-FID $\downarrow$  & IS $\uparrow$
                                            & Precision $\uparrow$ & Recall $\uparrow$  & F1-score $\uparrow$ \\
     \addlinespace[4pt] 
     \hline
     \addlinespace[4pt] 
     ImageBART~\cite{esser2021imagebart}    & 10.81 & -
                                            & 4.49 & - & - & -\\
     StyleGAN2~\cite{karras2020analyzing}    & \textbf{4.16} & $2.76^\dagger$
                                            & - & 0.71 & 0.46 & 0.558\\
     LDM~\cite{rombach2022high}             & 4.98 & - 
                                            & 4.50 & 0.73 & 0.50 & \textbf{0.593}\\
     LDM*~\cite{rombach2022high}             & 9.47 & 3.21 
                                            & 4.47 & 0.64 & \textbf{0.54} & 0.585 \\
     Ours                                   & 6.91 & \textbf{2.66} 
                                            & \textbf{4.56} & \textbf{0.80} & 0.45 & 0.574\\
     \addlinespace[4pt]
     \thickhline
    \end{tabular}
    
}
     \vspace{5pt}
     \caption{Quantitative results of generation on FFHQ dataset~\cite{karras2019style}. * denotes the results calculated using the checkpoint released by LDM authors on Github. The CLIP-FID ($^\dagger$) of StyleGAN2 is reported by another work~\cite{kynkaanniemi2022role}.}
    \label{tab:ffhq_quantitative_results}
\end{table}

%% file: tables/reconstruction_with_noise_table.tex
\begin{table}[h]
\centering
\resizebox{0.48\textwidth}{!}{
\tiny
\begin{tabular}{ccccc}
     \thickhline
     \addlinespace[2pt] 
     Method                                 &  Code Size & LPIPS$\downarrow$ & SSIM $\uparrow$ & PSNR $\uparrow$ \\ \hline
     \addlinespace[2pt] 
     \hline
     \addlinespace[2pt]
     Ours (w/o noise)                       &$32\times 32 \times 4$ & 0.12$\pm$0.07 & 0.67$\pm$0.21 & 22.29$\pm$6.14 \\
     Ours (with noise)                      &$32\times 32 \times 4$ & \textbf{0.11$\pm$0.07} & \textbf{0.68$\pm$0.20} & \textbf{22.45$\pm$5.98} \\
     \addlinespace[2pt]
     \thickhline
\end{tabular}}
 \vspace{4pt}
 \caption{Reconstruction metrics with vs.~without noise perturbation on the $\hat{z}_{i,j}^s$, evaluated on the validation split of LSUN-Church dataset~\cite{yu2015lsun} after training after 800K images.}
  \label{tab:noise_perturb_recon}

\end{table}

%% file: 05_conclusion.tex
\section{Conclusion, Limitation, and Future work}
\label{sec:future}
We introduced \abbrv, a novel bi-level feature representation that leverages a two-step process, where images are initially encoded into continuous key code blocks and then are employed to retrieve features from groups of multi-resolution feature grids. 
The key codes in \abbrv\ possess desired properties such as continuity, and relaxed data precision that help produce high-quality reconstruction and generation results. 
Our experimental evaluations showcase that our approach outperforms or achieves comparable reconstruction results to the popular Vector Quantization (VQ) methods while employing a smaller and more efficient decoder.
Furthermore, the generated results obtained by training a diffusion model on our key codes exhibit high-fidelity and improved structures compared to previous methods.
As a limitation, \abbrv\ requires more storage compared to existing methods because our feature grids accommodate highly dense feature vectors. It is worthwhile to explore the optimal size of feature grids that strikes a balance between compactness and reconstruction/generation quality.
Also, it is worth investigating the potential of our method for diverse data types such as videos, scenes, and 3D objects. These data types often exhibit higher redundancy in spatial and temporal dimensions compared to 2D images and could benefit from our multi-resolution feature grids and key code tiling technique. 

%% file: supp.tex
\appendix
\renewcommand{\thesection}{\Alph{section}}
\setcounter{section}{0}

\section*{Supplementary Material}
In this supplementary material, we first provide more implementation details of autoencoding and diffusion (Sec~\ref{sec:more_implementation_defails}).  Sec~\ref{sec:computational_overhead},\ref{sec:memory_overhead} are dedicated to an in-depth discussion of the computational and memory overhead introduced by our bi-level representation. 
Subsequently, we give more theoretical and experimental analysis on our relaxed precision setting and high utilization of feature grids (Sec~\ref{sec:relaxed-precision},\ref{sec:high-hit-ratio}). Sec~\ref{sec:evaluation_details} discusses more details of generation evaluation and Sec~\ref{sec:more_FFHQ_results},\ref{sec:more_LSUN_church_results} show more generation results. Finally, more parameter studies (Sec~\ref{sec:parameter_studies}) and ablation studies (Sec~\ref{sec:compatibility_with_other_decoders},\ref{sec:compatibility_with_hash_tables},\ref{sec:reconstruction_on_multi-class_dataset}) are presented.

\section{More Implementation Details}
\label{sec:more_implementation_defails}

\subsection{Detailed Architecture of the Autoencoder}
\label{sec:detailed_AE}

For more details about the encoder architecture, please refer to the official implementation\footnote{\href{https://github.com/CompVis/taming-transformers/blob/master/taming/modules/diffusionmodules/model.py}{https://github.com/CompVis/taming-transformers/blob/master/taming/modules/diffusionmodules/model.py}} of VQGAN~\cite{esser2021taming}. We use the same implementation but set the attention resolution from 64 to $\min(H_z,W_z)$ and the unit channel size to $32$. The maximum resolution ($r_{max}$) of feature grids is 32 if $H_z,W_z<=32$ and $r_{max}=64$ when $H_z,W_z=64$. 

\begin{figure*}[htb]
    \centering
    \includegraphics[width=0.9\textwidth]{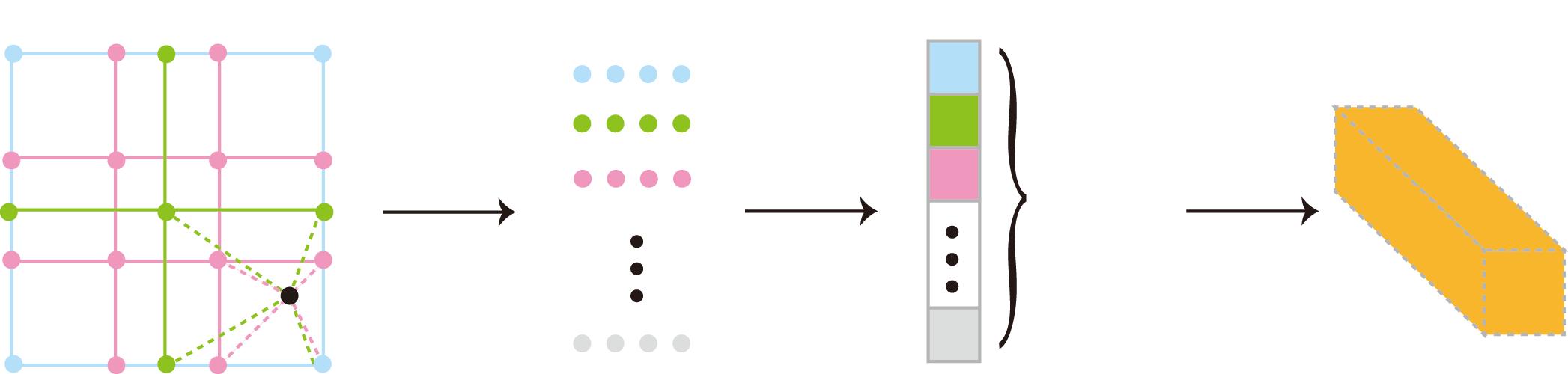}
    \begin{picture}(0,0)
    \put(-355,20){\makebox(0,0){$q^s_{i,j}$}}
    \put(-322,53){\makebox(0,0){$F_{r}$}}
    \put(-220,53){\makebox(0,0){$F_{inp}$}}
    \put(-98,53){\makebox(0,0){$F_{mlp}$}}
    \put(-140,45){\makebox(0,0){\tiny{$N_r\times C_v$}}}
    \put(-53,25){\tiny{\makebox(0,0){$C_d/N_h$}}}
    \put(-10,0){\makebox(0,0){$f^s_{i,j}$}}
  \end{picture}
    \caption{The feature retrieval and interpolation process inside each group of multi-resolution feature grids is illustrated for a 2D case. For each combined and sliced key code $q_{i,j}^s$, we would retrieve via $F_r$ and interpolate via $F_{inp}$ to get features across $N_r$ resolutions. Each resolution will give 4 nodes to be linearly interpolated and each node has dimension $C_v$. The feature vectors from all the resolutions will be concatenated as a vector of length $N_r\times C_v$. The concatenated feature vector will go through an MLP $F_{mlp}$ and be projected to be $f_{i,j}^s$ of length $C_d/N_h$ as part of the feature block $f$ to the decoder.}
    \label{fig:multi-resultion_feature_grids}
\end{figure*}

In \autoref{fig:multi-resultion_feature_grids}, we also illustrate the process of feature retrieval and interpolation inside the multi-resolution feature grids by each slice of key code $q_{i,j}^s$. 
For each level $l$ resolution of the feature grid, we set the solution as $r_l$. The surrounding grid corners $x$ of $q$ are obtained by $\lfloor q_{i,j}^sr_l \rfloor$ and $\lceil q_{i,j}^sr_l \rceil$, where $r_l=r_{min}b^l, b=\exp(\frac{\ln r_{max} - \ln r_{min}}{L-1}), L=16$ as the total number of resolutions following the setting of InstantNGP~\cite{muller2022instant}. The ultimate index of in the feature grid on the current resolution is:

\begin{equation}
    \text{index} = \big\{\sum^{C_{key}+2}_k{x_k\tau_k}\big\}\ \text{mod}\ T,
\end{equation}
where $\tau_k$ is the stride number of dimension $k$ and $T$ is the maximum allowed entry number of feature grids (in our default setting $C_{key}=1,T=2^{18}$). The total number of indices or corners is $2^{C_{key} + 2}$ and those retrieved values will be linearly interpolated by $F_{inp}$. 

\subsection{Training Details of the Autoencoder }
When training the autoencoder, we use the Adam optimizer~\cite{kingma2014adam} with learning rate $2e^{-3}$, $\beta_1=0,\beta_2=0.99$ for updating the encoder, decoder, feature grids and discriminator. We assign $\epsilon=1e^{-8}$ to the Adam optimizer of the discriminator and $\epsilon=1e^{-15}$ to the Adam optimizer of the encoder, decoder, and feature grids for quickly updating the entries inside auxiliary data structures~\cite{muller2022instant}.

\subsection{Loss Weights of Training Autoencoder}
We set the weight of $L_2$ loss as 20 and the weight of $L_{percep}$ as 5. The weight on gradient penalty is 4 across all experiments. We did not try to deliberately adjust this setting and not experience any training instability caused by the weight setting in the reconstruction stage.

\subsection{KL Regularization on Key Codes}
In addition to relaxed precision setting ($\epsilon_0$), we also tried KL-reg key codes obtained by $z=\text{sigmoid}(\mu + \sigma * \epsilon)$, $(\mu, \sigma)=E(x)$ where $\epsilon \sim \mathcal{N}(0,I)$, $\mu,\sigma \in \mathbb{R}^{H_z\times W_z \times C_z}$ and $E$ is the encoder. A KL divergence loss with weight $1e^{-6}$ (a typical weight used in previous KL regularized latents~\cite{rombach2022high}) is added to the autoencoder training stage.

\subsection{Detailed Architecture of Diffusion UNet}
Following the design of LDM~\cite{rombach2022high}, we use a UNet with channel multiplier of $(1,2,3,4)$, 2 residual blocks in each resolution and attention at resolutions $32,16,8$.

\subsection{Diffusion Model Training Details}
The training batch size of diffusion model is 64 and the learning rate is $8e^{-4}$ with linear learning rate warm-up at the first $1e^4$ training iterations and cosine learning rate decay for $1e^6$ iterations. We utilize $2\times$ Nvidia A100 GPUs (40GB memory per GPU) for 7 days and 400k iterations. Note that the noise perturbation $\epsilon_0$ is only added during autoencoder training and not added during the diffusion training. As for the two noise schedulers we adopt, we show the logSNR-t plot of the two noise schedulers in~\autoref{fig:logsnr_minsnr}. The parameter of the cosine variant is $\text{start=0.2},\text{end=1},\tau=1.5$. For more details on the cosine scheduler, please refer to Algorithm 1 of Chen 2023~\cite{chen2023importance}.

\section{Computational Overhead}
\label{sec:computational_overhead}
The additional computation cost measured by multiplication times introduced by our bi-level representation is:
\begin{equation}
\small
\begin{split}
\text{FLOPS}_{\text{\abbrv}} = & \underbrace{\frac{H_dW_dC_z}{C_{key}} * 2^{C_{key}+2} * C_vN_r}_{\text{linear interpolation}} + \\
 &\underbrace{H_dW_d * \big(2(C_vN_r)^2  + \frac{C_vN_rC_dC_{key}}{C_z}\big) * \frac{C_z}{C_{key}}}_{\text{MLP layers}}.
\end{split}
\end{equation}
Please refer to Table 1 of the main manuscript to recheck the meaning of the symbols. From the above equation, we can see $\text{FLOPS}_{\text{\abbrv}} \propto H_d,W_d,C_z.$ In Table 3 of our main manuscript, we can see that as $C_z$ increases, the overall FLOPs slightly increase (where we fix tiled key code size all to $(H_d,W_d)=(64,64)$). The computational cost of feature grids is \textbf{two orders of magnitude less} than it of the convolutional layers in the decoder, namely \textbf{0.27 GFLOPs} \textit{vs.} \textbf{63.40 GFLOPs} in total, when $(H_d,W_d,C_z)=(64,64,4)$. And among the 0.27 GFlops overhead, 95\% of it comes from MLP layers.

Moreover, we can see that computational cost is independent of the total number of entries in feature grids and it is only related to the number of points it queries during each forward process (namely $H_d\times W_d \times C_z$). In contrast, the computational cost of a convolutional layer grows with its kernel size.

\section{Memory Overhead}
\label{sec:memory_overhead}
The total memory cost introduced by our bi-level representation during training/inference is determined by the following operations: (1) loading auxiliary data structures into memory, (2) recording the feature vectors that encompass key codes, and (3) storing the gradients of these feature vectors:
\begin{itemize}
    \item The static memory footprint for each set of auxiliary data structures with maximum $2^{18}$ entries for each level of resolution and 4 float numbers of each entry is 18.64MB.
    \item We measure the training memory footprint by wrapping the whole feature retrieval module in \textit{torch.profile}. We measure the number through 10 iterations during training when $(H_d,W_d,C_z)=(64,64,4)$.
    \item Benefiting from our small overhead of memory footprint of the feature grids and the small decoder, we are able to train the reconstruction stage with batch size of \textbf{12} in our single 24 GB Nvidia Titan X GPU while we can only train VQGAN with batch size of \textbf{4} on that GPU. Note that we do not use any half/mixed-precision computation in the training of our implementation.
\end{itemize}
In conclusion, introducing feature grids will not bring significant memory cost (less than 100 MB for batch size of 1); a small number compared to the volume of memory that common neural networks consume during training.

\section{More Analysis of Relaxed Precision}
\label{sec:relaxed-precision}
In the diffusion inversion process, we aim to sample data points from the target distribition $p(x)$. Let us set the total inversion steps to be $T$ and consider the diffusion inversion process in the formulation of Langevin dynamics~\cite{song2019generative}:
\begin{equation}
    \tilde{x}_t = \tilde{x}_{t-1} + \frac{\epsilon}{2}\nabla_x \log p(\tilde{x}_{t-1})+\sqrt{\epsilon}z_t
\end{equation}
where $\tilde{x}_0$ belongs to arbitrary prior distribution, $z_t \sim \mathcal{N}(0,I)$ and $\nabla_x\log(\tilde{x}_{t-1})$ is the estimated score function of $\tilde{x}_{t-1}$. The distribution of $\tilde{x}_T$ will approach $p(x)$ when $T \rightarrow \infty,\epsilon \rightarrow 0$. In real world situations, this criterion cannot be met strictly. Instead, limited diffusion time steps and various noise scheduling strategies are adopted~\cite{ho2020denoising,nichol2021improved,chen2023importance}. Moreover, with a limited capacity of our diffusion UNet, the score function estimation would not be perfect either. As a result, there would be an accumulated error conforming to a Gaussian distribution between every timestep of predicted $\tilde{x}_t$ and the ground truth $x_t$~\cite{li2023alleviating,zhang2018unreasonable}, namely the exposure bias.
All in all, we are unable to approach arbitrary target data precision in the actual diffusion model setting.


Hence, in our setting, we add a certain noise perturbation $\epsilon_0$ that is based on the maximum resolution of feature grids to our continuous key codes to construct a relaxed and controlled precision. 

\begin{figure}
    \centering
    \includegraphics[width=0.3\textwidth]{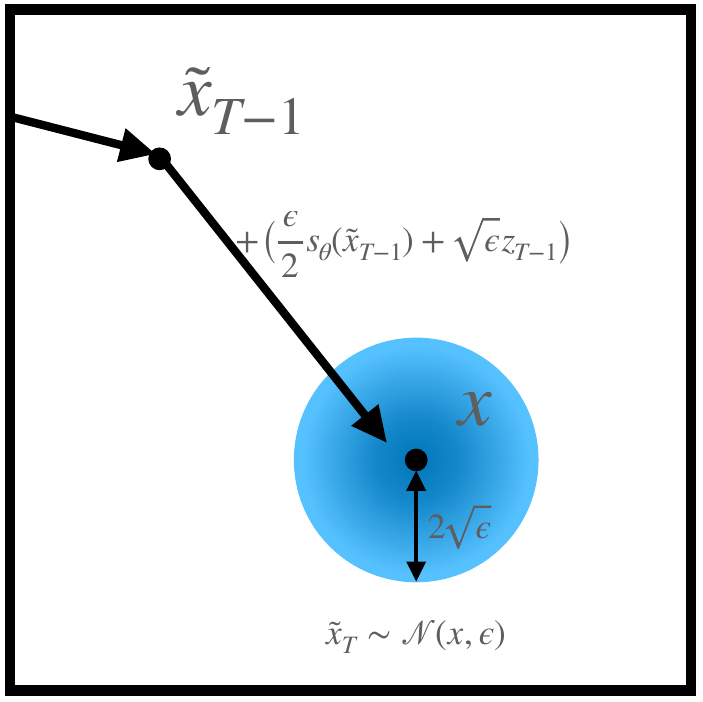}
    \caption{The diffusion inversion process with limited precision.}
    \label{fig:diffusion_path}
\end{figure}

\begin{figure*}
    \centering
    \includegraphics[width=0.95\textwidth]{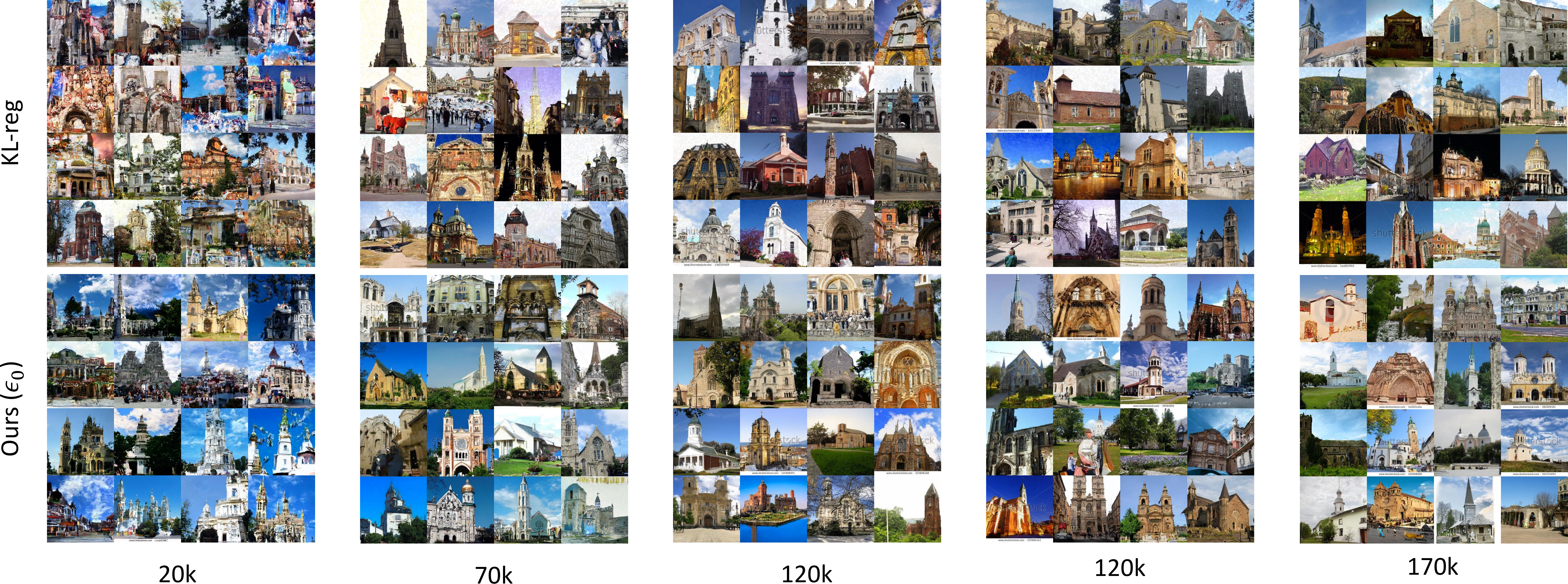}
    \caption{Random samples during training at different training stages (number of iterations) by KL-reg and our relaxed precision under the variant v2 of cosine noise scheduler. The spot artifacts persist in KL-reg while the artifacts do not appear in our relaxed precision setting. And at 170k iterations, our relaxed precision setting ($\epsilon_0$) gets 5k FID 8.47 vs KL-reg 9.64.}
    \label{fig:samples_during_training}
\end{figure*}

\begin{figure}
    \centering
    \includegraphics[width=0.5\textwidth]{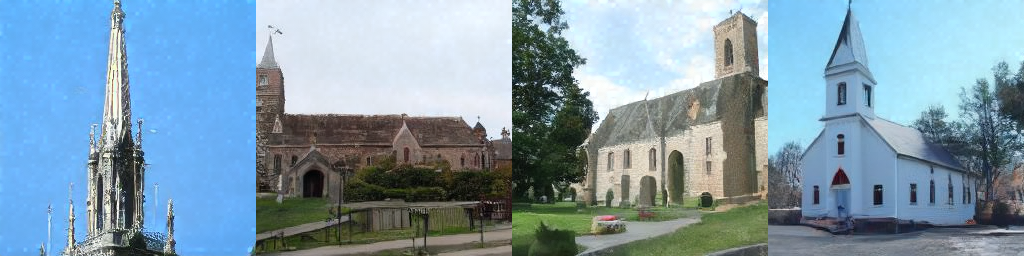}
    \caption{The irregular spot artifacts in KL-reg results with cosine variant noise scheduler. Zoom into the sky to look more clearly.}
    \label{fig:kl_reg_cosine_variant_artifacts}
\end{figure}

\begin{figure*}
    \centering
    \includegraphics[width=1.0\textwidth]{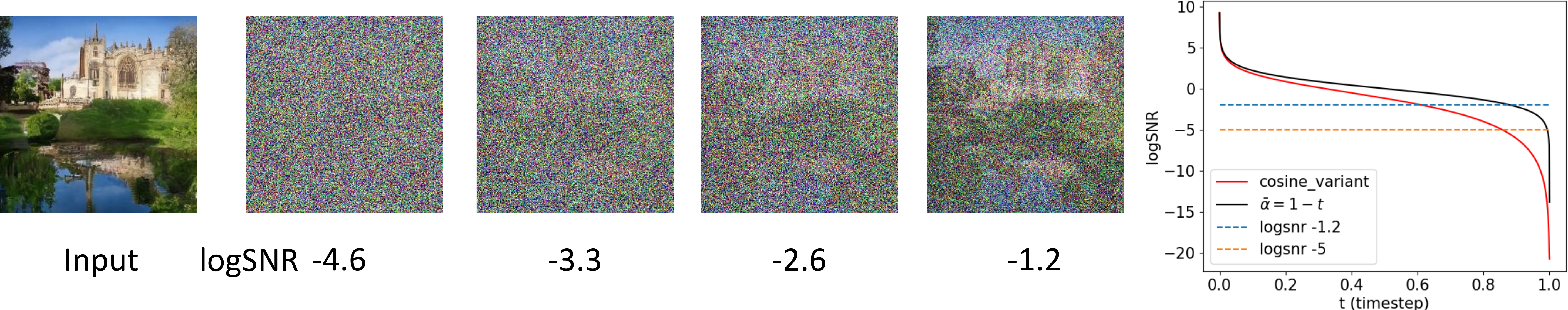}
    \caption{The lower SNR phase of diffusion process constructs the coarse shape of the generated result. $\bar{\alpha}=1-t$ noise scheduler contributes much \am{fewer} steps on lower SNR phase than the cosine variant one, resulting in worse mode coverage and FID scores.}
    \label{fig:lower_logsnr_show}
\end{figure*}

\begin{figure}[htbp]
    \centering
    \includegraphics[width=0.5\textwidth]{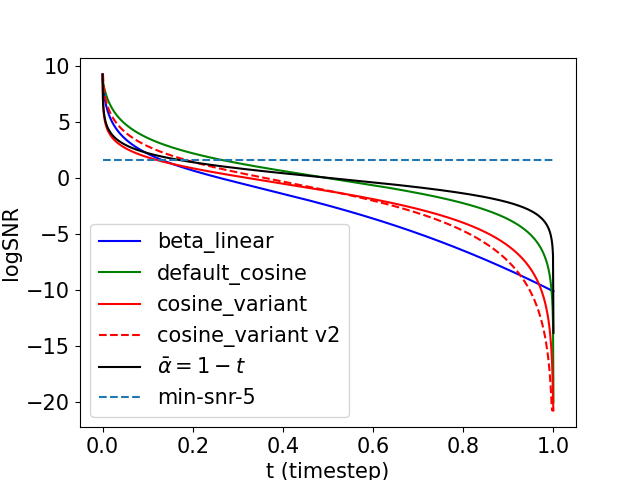}
    \caption{The logSNR-t plot with different noise schedulers. The dashed line here represents the min-snr-5 threshold. Through min-snr weighting strategy, earlier timesteps with bigger SNR than the threshold will be put with a smaller weight. The beta-linear noise scheduler from the original DDPM paper~\cite{ho2020denoising} is drawn here for reference.}
    \label{fig:logsnr_minsnr}
\end{figure}

\subsection{KL-reg vs. Relaxed Precision}
As we have shown in Section 4 of the main text, when applying diffusion models on KL-regularized code using cosine noise scheduler~\cite{nichol2021improved} with min-snr weighting strategy~\cite{hang2023efficient}, the generated results show spot artifacts. To further understand this, let us revisit  min-snr weighting strategy~\cite{hang2023efficient} first. Min-snr is a weighting strategy of $\omega_t = \min \{\text{SNR}(t), \gamma\}$, $\gamma$ is a hyper-parameter and we use the default value 5 from the original min-snr paper. As shown in \autoref{fig:logsnr_minsnr}, this weighting strategy pushes the diffusion model to avoid focusing too much on small noise levels (initial timesteps) and helps it to be trained faster. While our relaxed precision method is compatible with Min-snr, the KL-regularized key codes cannot converge to precise neighborhoods that avoid the artifacts. In fact, even with cosine variant noise scheduler which is not combined with min-snr weighting strategy, the KL-reg results still suffer from some spot artifacts as shown in \autoref{fig:kl_reg_cosine_variant_artifacts}. In the \autoref{fig:samples_during_training}, we show randomly sampled images during training with KL-reg and our relaxed precision setting under the variant v2 of the cosine noise scheduler. From this figure, we demonstrate that our relaxed precision setting can be adapted to different noise schedulers. Moreover, quantitatively our relaxed precision setting gets 5k FID score 8.47 vs KL-reg 9.64 at 170k iterations.


To further strengthen our argument, we collect the statistics of $\sigma$ in KL-regularized key codes of 5,000 images in training set $\sigma=8.95\times 10 ^ {-5}\pm 8.0 \times 10 ^ {-4}, (\mu, \sigma)= E(x)$, $E$ is the encoder and $x$ is the input image. Compared with our relaxed precision setting ($\epsilon_0$), $\sigma = \frac{1}{(4r_{max})} \approx 3.9 \times 10^{-3}, r_{max}=64$, ours has much lower precision. In fact, the data precision of KL-regularized key codes is even higher because there is one additional sigmoid transformation after the reparameterization trick.

\subsection{More noise scheduler}
We also test $\bar{\alpha}=1-t$ noise scheduler~\cite{chen2023importance}($q(\textbf{x}_t
|\textbf{x}_0)=\mathcal{N}(\textbf{x}_t;\sqrt{\bar{\alpha_t}},(1 - \bar{\alpha_t})\textbf{I})$, $\textbf{x}_0$ here is the original data distribution.)  on KL-reg and our relaxed precision setting on our key codes. As we can see in \autoref{fig:logsnr_minsnr}, this $\bar{\alpha}=1-t$ noise scheduler focuses much less on lower SNR than others while the lower SNR phase of diffusion constructs the coarse shape of the image (see \autoref{fig:lower_logsnr_show} for more visualization). Under this noise scheduler, we do not observe any apparent spot artifacts in either KL-reg or relaxed precision settings. But at the same time, the 5k FID scores for the two settings are 9.80 vs 9.33, which are much higher than other noise schedulers due to the lack of diversity in generated outputs. After all, our relaxed precision method still outperforms the KL-reg key codes.

\section{More Analysis of the High Utilization of Feature Grids}
\label{sec:high-hit-ratio}
A classic vector-quantization training target is~\cite{yu2021vector}:
\begin{equation}
    L_{\text{VQ}} = \|\text{sg}[z_e(x)]-e\|^2_2 + \beta \|z_e(x)-\text{sg}[e]\|_2^2
\end{equation}
where $z_e(x)$ is the code obtained by encoding the input image, $e$ is the code stored in the codebook, and $\text{sg}$ is the stop gradient operator.

One of the key features of the Vector Quantization (VQ) process is the clustering-like behaviour where the index of $i$ of the code $e$ in the codebook looked up by $z_e$ is determined by $i=\arg \min_j\|z_e(x)-e_j\|_2^2$. One immediate issue that arises is the \textit{clustering of high dimensional data}. The typical dimension of the encoded features is in the hundreds and the distance measurement becomes less precise when dimension increases~\cite{assent2012clustering}. As shown in Table 4 of ViT-VQGAN~\cite{yu2021vector}, when the latent dimension increases from 4 to 256, the codebook usage decreases from 96\% to 4\%.

Instead of entangling the index selection with discontinuous, clustering-like value comparison process, we disentangle index (key code)/value (features) learning by directly encoding images to key codes and update them continuously with well-defined gradient flows.



\section{Evaluation Details}
\label{sec:evaluation_details}
\textbf{FID and CLIP-FID} In accordance with recent research~\cite{kynkaanniemi2022role}, it has been found that the Fréchet Inception Distance (FID) occasionally exhibits disparities with human evaluative judgments. The substantial improvement of FID may occur without an actual improvement in generation quality, owing to intentional or accidental distortions aligned with the feature space of ImageNet classes. This is attributed to the fact that the FID feature extractor is trained on the ImageNet classification task. Therefore, the utilization of CLIP-FID is recommended as an additional metric alongside FID. The CLIP-FID metric employs the pre-trained CLIP~\cite{radford2021learning} feature extractor for feature extraction. Table 4 in the main text shows that Projected GAN~\cite{sauer2021projected} attains the lowest FID and highest Inception score, coupled with the highest CLIP-FID score among the methods reported. This outcome coincides with the aforementioned findings in the study~\cite{kynkaanniemi2022role}. 

\textbf{Libraries} Our evaluation scripts are built on widely-adopted open source Python libraries including: 
 \textit{PerceptualSimilarity}\footnote{\href{https://github.com/richzhang/PerceptualSimilarity}{https://github.com/richzhang/PerceptualSimilarity}} for LPIPS, \textit{torchmetrics}\footnote{\href{https://torchmetrics.readthedocs.io/en/stable/}{https://torchmetrics.readthedocs.io/en/stable/}} for PSNR and SSIM, \textit{torch-fidelity}\footnote{\href{https://github.com/toshas/torch-fidelity}{https://github.com/toshas/torch-fidelity}} for FID, \textit{clean-fid}\footnote{\href{https://github.com/GaParmar/clean-fid}{https://github.com/GaParmar/clean-fid}} for CLIP-FID and a 3rd-party repo\footnote{\href{https://github.com/youngjung/improved-precision-and-recall-metric-pytorch}{https://github.com/youngjung/improved-precision-and-recall-metric-pytorch}} for the precision and recall score. All the evaluation scripts are included in our anonymous \href{https://anonymous.4open.science/r/brics_code_release-D0F5}{code release}.


\section{More Results on FFHQ~\cite{karras2019style} Dataset}
\label{sec:more_FFHQ_results}

\begin{figure*}
    \centering
    \includegraphics[width=0.8\textwidth]{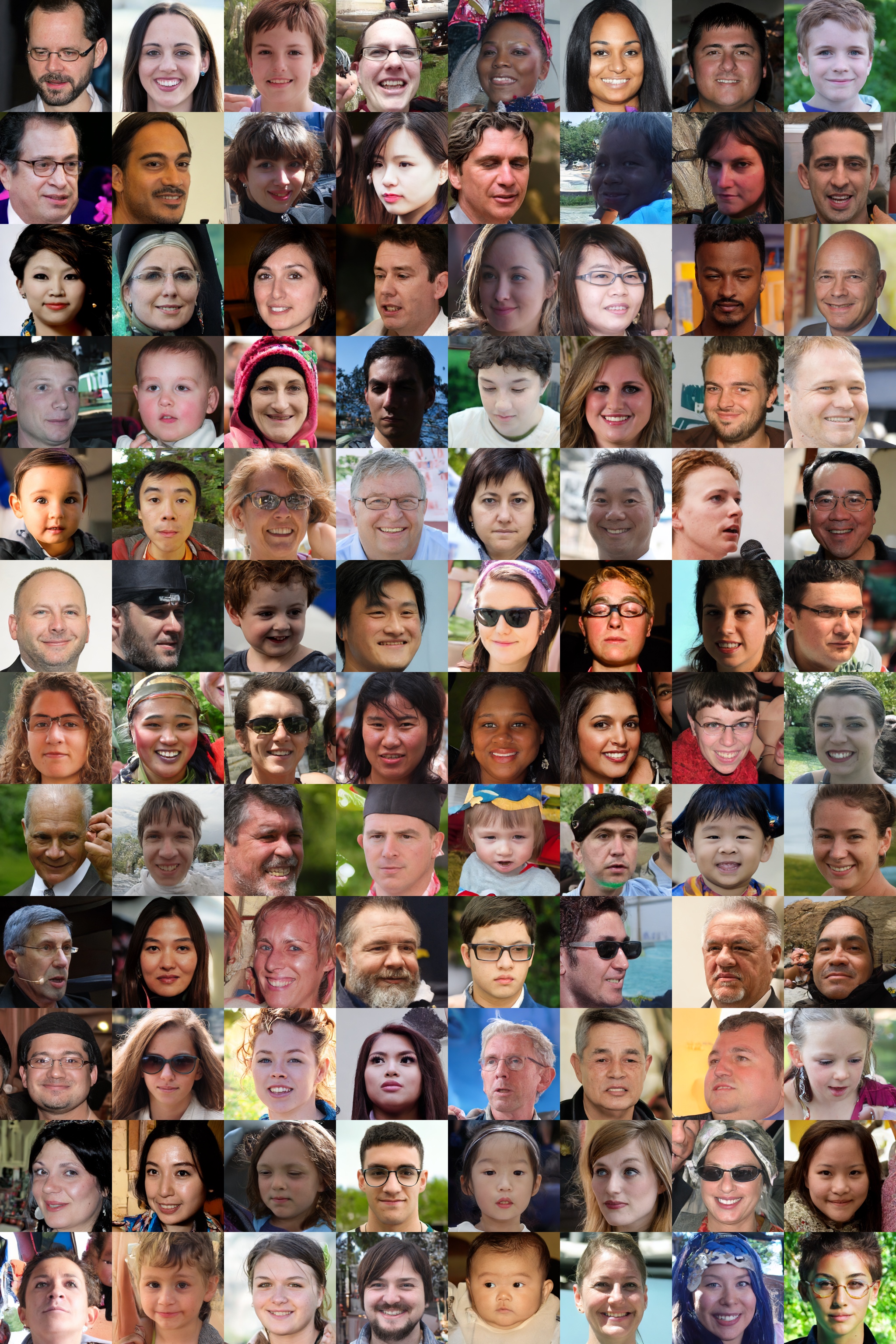}
    \caption{Uncurated samples of our generation results on FFHQ $256\times256$.}
    \label{fig:uncurated_samples_ffhq}
\end{figure*}
\begin{figure*}
    \centering
    \includegraphics[width=0.8\textwidth]{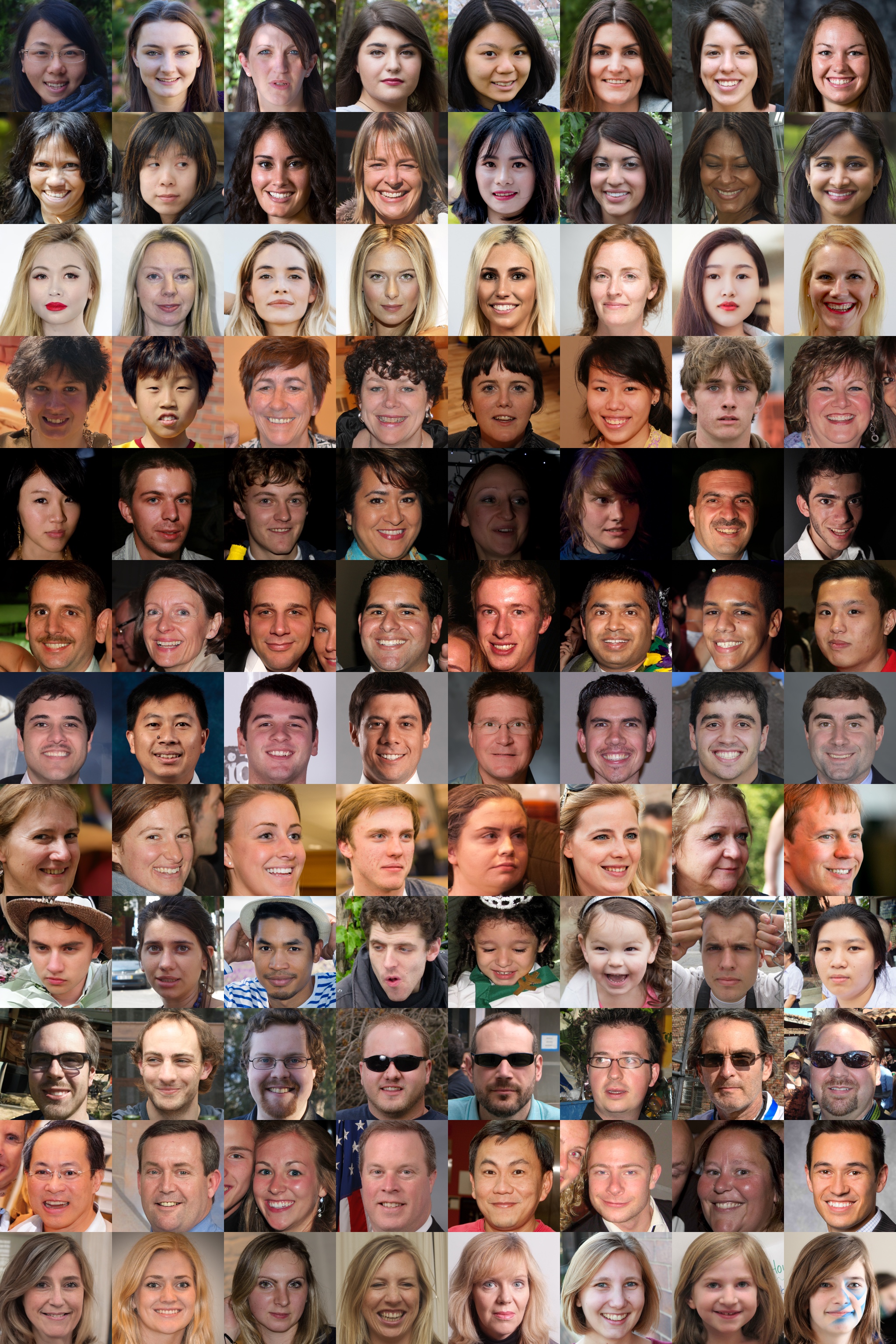}
    \caption{Nearest (by LPIPS~\cite{zhang2018unreasonable}) neighbours of uncurated samples of our generation results on FFHQ $256\times256$. The \textbf{leftmost} images in each row are the generated samples and the rest images in each row are nearest neighbours in training dataset.}
    \label{fig:uncurated_nearest_samples_ffhq}
\end{figure*}

We showcase more visual results in \autoref{fig:uncurated_samples_ffhq} from FFHQ.  From the uncurated samples, we can see that the diffusion model trained on key codes is able to generate high-fidelity human faces with diverse races, ages, hair styles, poses, accessories and so on. Moreover, we also provide more nearest neighbour results in \autoref{fig:uncurated_nearest_samples_ffhq} to restate that even with much higher precision score, our generated results are unique samples and not mere retrievals from the training set.

\section{More Results on LSUN-Church~\cite{yu2015lsun} Dataset}
\label{sec:more_LSUN_church_results}

\begin{figure*}
    \centering
    \includegraphics[width=0.8\textwidth]{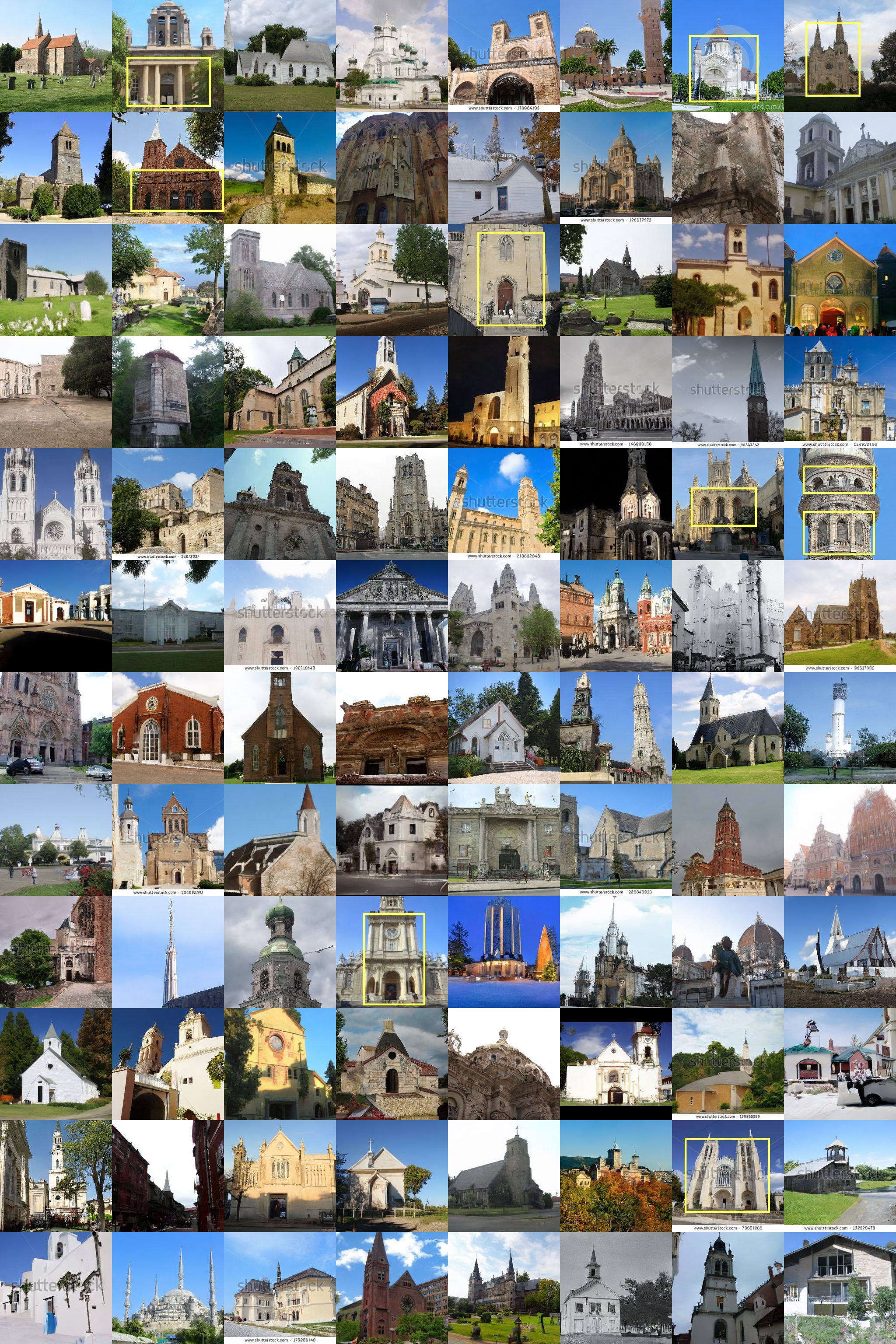}
    \caption{Uncurated samples of our generated results on LSUN-Church $256\times256$. \textcolor{yellow}{Yellow} squares highlight some of our structured and symmetric results.}
    \label{fig:uncurated_samples_lsun_church}
\end{figure*}

\begin{figure*}
    \centering
    \includegraphics[width=0.8\textwidth]{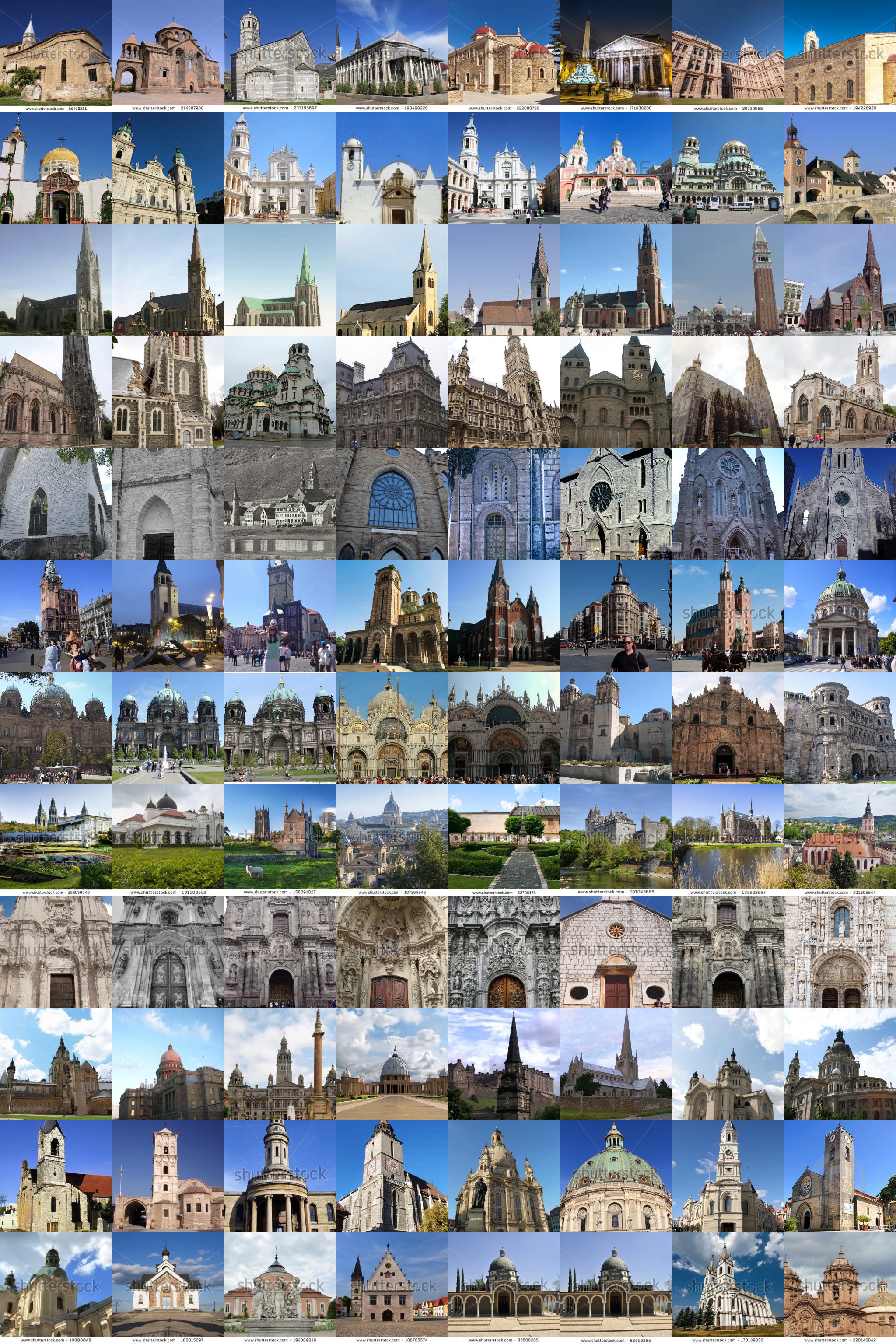}
    \caption{Nearest (by LPIPS~\cite{zhang2018unreasonable}) neighbours of uncurated samples of our generation results on LSUN-Church $256\times256$. The \textbf{leftmost} images in each row are the generated samples and the rest images in each row are nearest neighbours in training dataset.}
    \label{fig:uncurated_nearest_samples_lsun_church}
\end{figure*}

\begin{figure*}[h]
    \centering
    \includegraphics[width=0.8\textwidth]{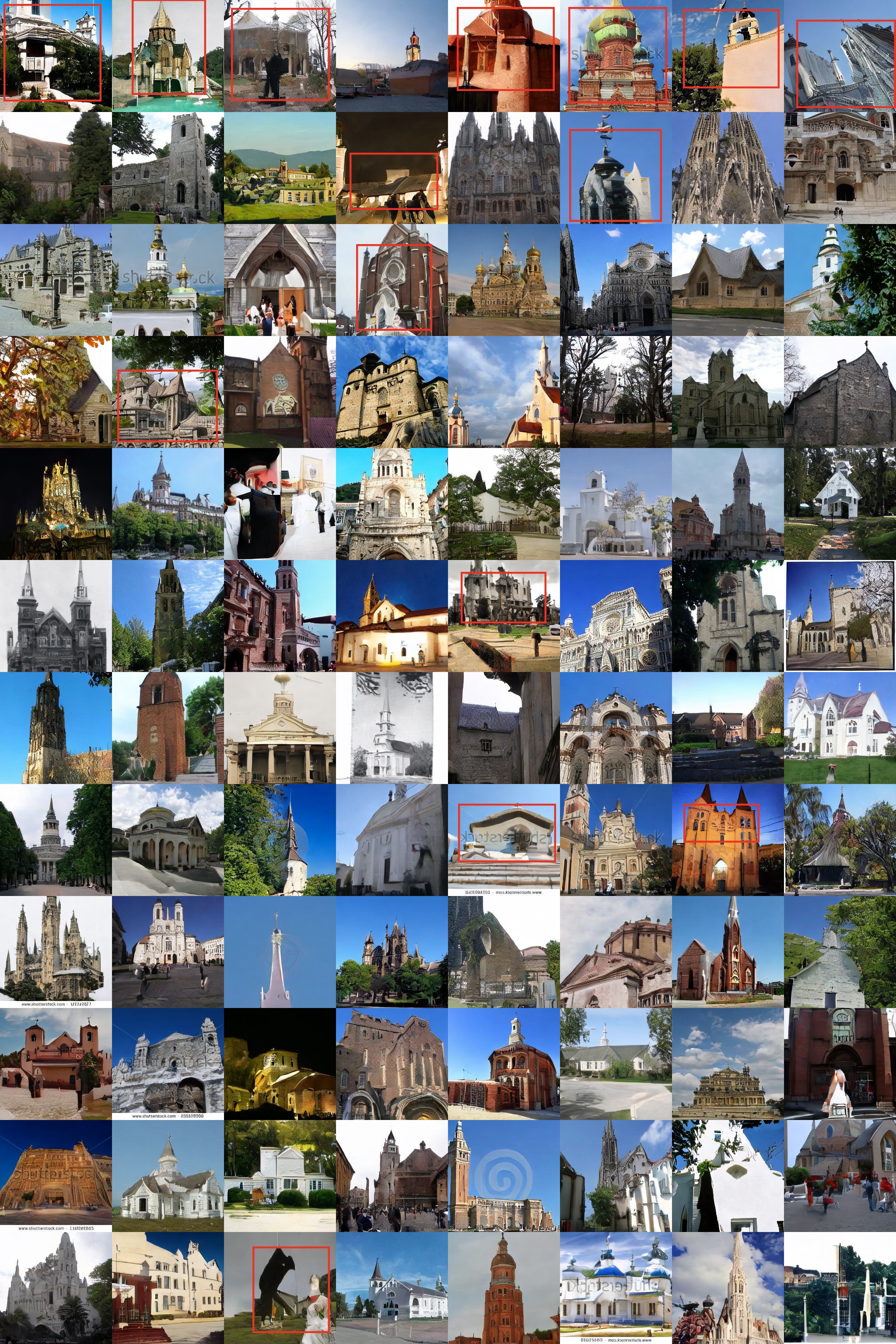}
    \caption{Uncurated samples of LDM~\cite{rombach2022high} generation results on LSUN-Church $256\times256$. \textcolor{red}{Red} squares highlight some structural issues in LDM results.}
    \label{fig:uncurated_ldm_churches}
\end{figure*}

 
 We showcase more visual results \autoref{fig:uncurated_samples_lsun_church} from LSUN-Church. As we can see from the uncurated samples, our method \textit{consistently} generates diverse church buildings with good overall structures and details, coinciding with our high precision score. We also provide more nearest neighbour results in \autoref{fig:uncurated_nearest_samples_lsun_church}. In addition, we show more uncurated samples from LDM~\cite{rombach2022high} in \autoref{fig:uncurated_ldm_churches} to highlight their difference with our results. Generated images from LDM  sometimes lack symmetry, and also have bent columns and unstructured windows. We use \textcolor{red}{red} squares to highlight such distortions of LDM results in \autoref{fig:uncurated_ldm_churches} and \textcolor{yellow}{yellow} squares to highlight our structured and symmetric results in \autoref{fig:uncurated_samples_lsun_church}.


\section{Parameter Studies}
\label{sec:parameter_studies}
\input{tables/c_key_ablation_study}
\input{tables/hdwd_ablation_study}

\subsection{The Setting of $C_{key}$}

In this section, we discuss how the value of $C_{key}$, i.e., the length of a key unit, will affect the performance and complexity of our method.
We tested $C_{key} \in \{1,2\}$ with the code size of $64\times64\times4$, and kept all other hyper-parameters fixed. As shown in \autoref{tab:c_key_ablation}, increasing $C_{key}$ from $1$ to $2$ leads to a marginal improvement in reconstruction metrics on the LSUN-Church~\cite{yu2015lsun} validation split. However, it brings a significant increase of the size of feature grids, from 11.4M to 30.5M. The increase is due to the fact that, for each resolution $r$ of the feature grids, the total number of parameters is $\min(r^{2+C_{key}}, 2^{18})$, where $2^{18}$ is the maximum allowed entry number for each feature grid. Therefore, there exists a trade-off between performance and the amount of feature grid parameters.
We choose $C_{key} = 1$ as the default setting.

\subsection{The Setting of $H_d$ and $W_d$}

We also investigated the impact of adjusting the values of $(H_d,W_d)$, i.e., the resolution of feature blocks that are sent to the decoder. We tested $(H_d,W_d) \in \{(64,64), (32,32)\}$ with the code size of $32\times32\times8$. When $(H_d,W_d) = (64, 64)$, the key code will be tiled from $32\times32\times8$ to $64\times64\times8$. When $(H_d,W_d) = (32, 32)$, an additional convolution block is required to bring up the resolution from $32$ to $64$ in the decoder, yielding more training parameters and computation. As shown in \autoref{tab:hdwd_ablation_study}, setting $(H_d,W_d)=(64,64)$ both reduces the computational complexity and offers slightly better reconstruction performance. Therefore, we choose $(H_d,W_d)=(64,64)$ as the default setting.

\input{tables/compatability_comparison}

\section{Compatibility with Other Decoders}
\label{sec:compatibility_with_other_decoders}
To demonstrate the compatibility of our bi-level representation,  we conduct experiments on different decoder architectures. As \autoref{tab:decoder_comparison} shows, our bi-level representation is compatible with the decoders of VQGAN\cite{esser2021taming} and MoVQ\cite{zheng2022movq}. Moreover, our decoder with our bi-level representation outperforms theirs in LPIPS and PSNR metrics on LSUN-Church dataset. It is noted that MoVQ\cite{zheng2022movq} is computationally heavy (\textbf{100.40 GFlops} decoder) and memory intensive (\textbf{$\sim$10 GB more memory} during training than ours) spatially conditional normalization. The MoVQ implementation comes from a third-party implementation~\footnote{\href{https://github.com/ai-forever/Kandinsky-2}{https://github.com/ai-forever/Kandinsky-2}} since we did not find the official implementation. All in all, we tried our best to prove that our bi-level representation can be coupled with different decoder architectures and the possibility of improving the decoder architecture for better reconstruction and generation results is left for further study. 

\section{Compatibility with Hash Tables}
\label{sec:compatibility_with_hash_tables}
\begin{figure}[htbp]
    \centering
    \includegraphics[width=0.5\textwidth]{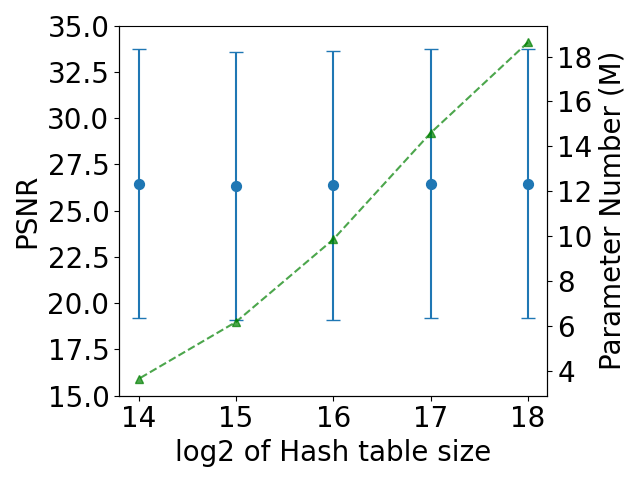}
    \caption{The PSNR metric of the reconstruction in LSUN-Church validation split and the parameter number of hash tables change with the maximum entry number for each level of hash table.}
    \label{fig:hash_table_size_change}
\end{figure}

When we change the regular spatial retrieval function $F_r$ to universal hashing function, the multi-resolution feature grids would naturally become multi-resolution hash tables. We have done a series of experiments on LSUN-Church dataset to show our method's compatibility with hash tables and possible ways to further compress the number of parameters in the auxiliary data structure by decreasing the maximum entry number for each level of hash table (from $2^{18}$ to $2^{14}$). We fix other parameters $H_z,W_z=64,r_{max}=64,H_d,W_d=64$. As we can see from \autoref{fig:hash_table_size_change}, the parameter number can be compressed one magnitude further when decreasing the maximum entry number parameter from $2^{18}$ to $2^{14}$ while maintaining the reconstruction performance in the perspective of PSNR metric. It is for future study to show what effect this kind of further compression would have on following diffusion model training.

\begin{figure}[h]
\centering
\includegraphics[width=0.5\textwidth]{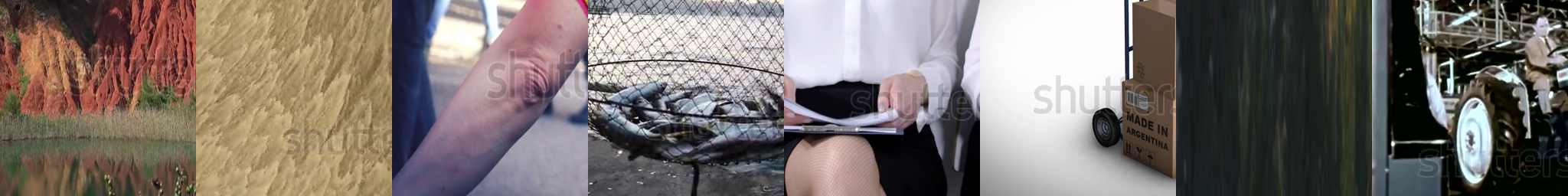}
\caption{Randomly selected samples from the training dataset of Webvid-Frames200k.}
\label{fig:webvid_frames_training}
\end{figure}

\begin{figure}[h]
\centering
\includegraphics[width=0.5\textwidth]{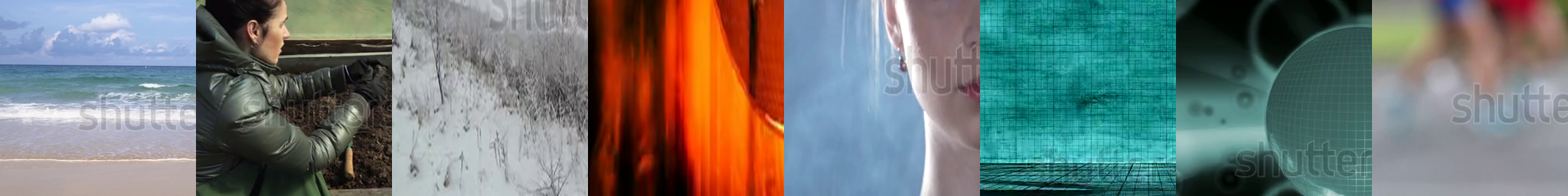}
\caption{Randomly selected samples from the test dataset of Webvid-Frames200k.}
\label{fig:webvid_frames_test}
\end{figure}

\section{Reconstruction on Multi-class Dataset}
\label{sec:reconstruction_on_multi-class_dataset}
\begin{figure}[t]
\centering
\includegraphics[width=0.5\textwidth]{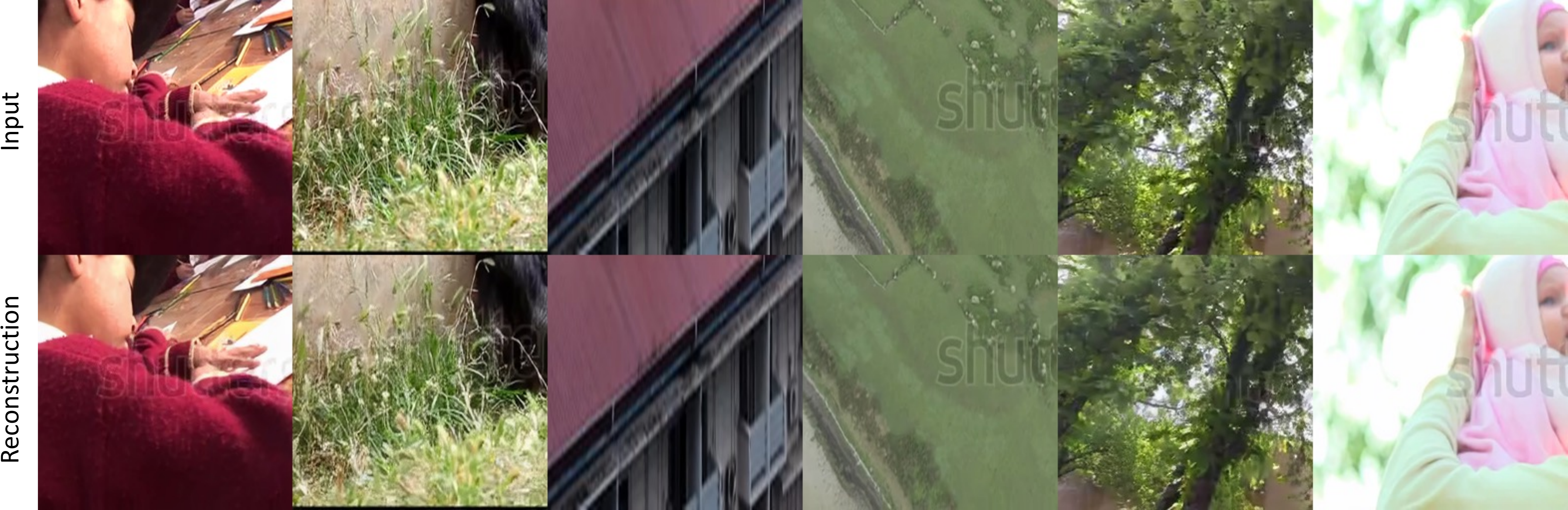}
\caption{Uncurated reconstruction results on test split of Webvid-Frames200k on the code size of $64\times64\times4$.}
\label{fig:webvid_frames_recon_results}
\end{figure}

We conduct the reconstruction task on a new dataset derived from the Webvid-10M dataset~\cite{bain2021frozen}. We randomly cropped 200k of images from random video clips of each video in the training split as our training dataset and 5k images from the validation split as our test dataset. This is a dataset with very high variety and we show some example images in \autoref{fig:webvid_frames_training} and \autoref{fig:webvid_frames_test}. We name this dataset \textbf{\textit{Webvid-Frames200k}}. The script for constructing this dataset has been included in the code release. Our model with the default setting achieves the values shown in \autoref{tab:web_vid_recon} and some uncurated reconstruction results are shown in \autoref{fig:webvid_frames_recon_results}. From this results, we demonstrate that our bi-level representation can also be adapted to datasets with very high variety.

\input{tables/webvid_recon_results_table}

\newpage

%% file: tables/c_key_ablation_study.tex
\begin{table*}[!htp]
    \centering 
    \resizebox{0.8\textwidth}{!}{
    \begin{tabular}{cccccccc}
     \thickhline
     \addlinespace[4pt] 
     Method                                 & Code size  & $C_{key}$ 
                                            & LPIPS$\downarrow$ & SSIM$\uparrow$ 
                                            & PSNR$\uparrow$ & \makecell{Decoder \\ \#Parameter} & \makecell{Decoder \\ GFlops} \\ 
     \addlinespace[4pt] 
     \hline
     \addlinespace[4pt] 
     Ours                                   & $64\times64\times4$ & 1
                                            & 0.0383 $\pm$ 0.0312 & 0.8438 $\pm$ 0.1295 & 26.34 $\pm$ 7.136& \textbf{18.3M$^*$ + 9.9M} & \textbf{63.41} \\
     Ours                                   & $64\times64\times4$ & 2
                                            & \textbf{0.0365 $\pm$ 0.0297} & \textbf{0.8480 $\pm$ 0.1282} & \textbf{26.47 $\pm$ 7.289}
                                            & 30.5M$^*$ + 9.9M & \textbf{63.41} \\
     \addlinespace[4pt]
     \thickhline
    \end{tabular}
    
}
     \vspace{5pt}
     \caption{Ablation study on $C_{key}$ on LSUN-Church~\cite{yu2015lsun}. $^*$ indicates the total number of parameters in feature grids.}
    \label{tab:c_key_ablation}
\end{table*}

%% file: tables/hdwd_ablation_study.tex
\begin{table*}[!tp]
    \centering 
    \resizebox{0.8\textwidth}{!}{
    \begin{tabular}{cccccccc}
     \thickhline
     \addlinespace[4pt] 
     Method                                 & Code size  & $H_d,W_d$ 
                                            & LPIPS$\downarrow$ & SSIM$\uparrow$ 
                                            & PSNR$\uparrow$ & \makecell{Decoder \\ \#Parameter} & \makecell{Decoder \\ GFlops} \\ 
     \addlinespace[4pt] 
     \hline
     \addlinespace[4pt] 
     Ours                                   & $32\times32\times8$ & $64,64$ 
                                            & \textbf{0.0697 $\pm$ 0.0516} & \textbf{0.7576 $\pm$ 0.1735} & \textbf{24.23 $\pm$ 6.622}& \textbf{21.2M$^*$ + 9.9M} & \textbf{63.54} \\
     Ours                                   & $32\times32\times8$ & $32,32$ 
                                            & 0.0690 $\pm$ 0.0505 & 0.7466 $\pm$ 0.1799 & 24.06 $\pm$ 6.713
                                            & 21.2M$^*$ + 35.1M & 85.14 \\
     \addlinespace[4pt]
     \thickhline
    \end{tabular}
    
}
     \vspace{5pt}
     \caption{Ablation study on $H_d,W_d$ on LSUN-Church~\cite{yu2015lsun}. An $^*$ indicates total total number of parameters in feature grids.}
    \label{tab:hdwd_ablation_study}
\end{table*}

%% file: tables/compatability_comparison.tex
\begin{table}[h]
\centering
\resizebox{0.48\textwidth}{!}{
\tiny
\begin{tabular}{cccc}
     \thickhline
     \addlinespace[2pt] 
     Decoder                                 &  LPIPS$\downarrow$ & SSIM $\uparrow$ & PSNR $\uparrow$ \\ \hline
     \addlinespace[2pt] 
     \hline
     \addlinespace[2pt]
     Ours                         & \textbf{0.0383} $\pm$ \textbf{0.0312} & 0.8438$\pm$0.1295 & \textbf{26.34} $\pm$ \textbf{7.136} \\
     VQGAN\cite{esser2021taming}  & 0.0535 $\pm$ 0.0400 & 0.8126$\pm$0.1457 & 25.27 $\pm$ 6.581 \\
     MoVQ\cite{zheng2022movq}  & 0.0396 $\pm$ 0.0327 & \textbf{0.8519} $\pm$ \textbf{0.1240} & 26.30 $\pm$ 7.102 \\
     \addlinespace[2pt]
     \thickhline
\end{tabular}}
 \vspace{4pt}
 \caption{Quantitive results on the validation split of LSUN-Church dataset for different decoder choices in the code size of $64\times64\times4$.}
  \label{tab:decoder_comparison}

\end{table}

%% file: tables/webvid_recon_results_table.tex
\begin{table}[h]
\centering
\resizebox{0.48\textwidth}{!}{
\tiny
\begin{tabular}{cccc}
     \thickhline
     \addlinespace[2pt] 
     Code Size                                 &  LPIPS$\downarrow$ & SSIM $\uparrow$ & PSNR $\uparrow$ \\ \hline
     \addlinespace[2pt] 
     \hline
     \addlinespace[2pt]
     32 $\times$ 32 $\times$ 4                  & 0.0769 $\pm$ 0.1109 & 0.8325 $\pm$ 0.2709 & 30.34 $\pm$ 10.73 \\
     64 $\times$ 64 $\times$ 4  & 0.0278 $\pm$ 0.0519 & 0.9242 $\pm$ 0.1321 & 34.61 $\pm$ 10.64 \\
     \addlinespace[2pt]
     \thickhline
\end{tabular}}
 \vspace{4pt}
 \caption{Quantitative results on the test split of Webvid-Frames200k dataset.}
  \label{tab:web_vid_recon}

\end{table}